\documentclass[times,twocolumn]{elsarticle}

\usepackage{medima}
\usepackage{framed,multirow}

\usepackage{amsmath,amssymb,amsfonts}
\usepackage{algorithmic}
\usepackage{latexsym}

\usepackage{url}
\usepackage{xcolor}

\usepackage{hyperref}

\definecolor{newcolor}{rgb}{.8,.349,.1}
\newcommand{\toadd}[1]{{#1}}

\begin{document}

\verso{Dang, Galati \textit{et~al.}}

\begin{frontmatter}

\title{Vessel-CAPTCHA: an efficient learning framework for vessel annotation and segmentation}

\author[1,2]{Vien Ngoc {Dang}\fnref{fn1}}
\author[1]{Francesco {Galati} \fnref{fn1}}
\fntext[fn1]{Joint first auhtorship.}
\author[3]{Rosa {Cortese}}
\author[1,4]{Giuseppe {Di Giacomo}}
\author[1,4]{Viola {Marconetto}}
\author[1]{Prateek {Mathur}}
\author[2] {Karim {Lekadir}}

\author[5]{Marco {Lorenzi}}
\author[6,3,7,8]{Ferran {Prados}}
\author[1]{Maria A. {Zuluaga}\corref{cor1}}

\address[1]{Data Science Department, EURECOM, Sophia Antipolis, France}
\address[2]{Artificial Intelligence in Medicine Lab, Facultat de Matemàtiques I Informàtica, Universitat de Barcelona, Spain}
\address[3]{Queen Square MS Centre, Dept of Neuroinflammation, UCL Queen Square Institute of Neurology, Faculty of Brain Sciences, University College London, UK}
\address[4]{Politecnico di Torino, Turin, Italy}
\address[5]{Université Côte d’Azur, Inria Sophia Antipolis, Epione Research Group, Valbonne, France}
\address[6]{Centre for Medical Image Computing, Dept of Medical Physics and Bioengineering, University College London, UK}
\address[7]{National Institute for Health Research, University College London Hospitals, Biomedical Research Centre, London, UK}
\address[8]{e-health Center, Universitat Oberta de Catalunya, Barcelona, Spain}


\begin{abstract}
Deep learning techniques for 3D brain vessel image segmentation have not been as successful as in the segmentation of other organs and tissues. This can be explained by two factors. First, deep learning techniques tend to show poor performances at the segmentation of relatively small objects compared to the size of the full image. Second, due to the complexity of vascular trees and the small size of vessels, it is challenging to obtain the amount of annotated training data typically needed by deep learning methods. 
To address these problems, we propose a novel annotation-efficient deep learning vessel segmentation framework. The framework avoids pixel-wise annotations, only requiring \toadd{weak} patch-level labels to discriminate between vessel and non-vessel 2D patches in the training set, in a setup similar to the CAPTCHAs used to differentiate humans from bots in web applications. The user-provided \toadd{weak} annotations are used for two tasks: 1) to synthesize pixel-wise \toadd{pseudo-}labels for vessels and background in each patch, which are used to train a segmentation network, and 2) to train a classifier network. The classifier network allows to generate additional weak patch labels, further reducing the annotation burden, and it acts as a noise filter for poor quality images. We use this framework for the segmentation of the cerebrovascular tree in Time-of-Flight angiography (TOF) and Susceptibility-Weighted Images (SWI). The results show that the framework achieves state-of-the-art accuracy, while reducing the annotation time by \toadd{$\sim$77\% w.r.t.} learning-based segmentation methods using pixel-wise labels for training.

\end{abstract}


\end{frontmatter}

\section{Introduction}
\label{sec:introduction}
The segmentation of the 3D brain vessel tree is a crucial task to the diagnosis, management, treatment and intervention of a wide range of conditions with a vast population-level impact \citep{who2018}. Due to the high complexity of the cerebrovascular tree, its automatic extraction is a challenging task. Despite decades of research  \citep{Lesage2009,Moccia2018}, the problem remains open.

With the advent of machine learning and, more precisely, deep learning techniques over the last decade \citep{Litjens2017,Lundervold2019}, image segmentation of organs, organs substructures, 
and lesions 
has reached state-of-the-art performance. This progress, however, has not been as fast in 3D brain vessel segmentation. Differently from the segmentation of other organs, there is no consolidated deep learning method which has reached human performance, and a vast majority of methods \toadd{\citep{Bernier2018,li2014,li2019,Morrison2018}} still rely on more classical techniques. 
This lag can be explained by two factors. First, deep learning techniques \toadd{often} assume that the object to segment occupies an important part of the image \citep{Deng2009,Shelhamer2017}. On the opposite, vessels are relatively small objects within a large image volume \citep{Livne2019,tetteh2020}. Secondly, deep learning techniques are well-known for being data greedy, as they require large annotated training datasets to avoid poor generalization. Due to the complexity of vascular trees and the small size of vessels, it is challenging to obtain sufficiently large high-quality annotated sets. 

This work presents a novel framework to address the challenges faced by deep learning-based 3D vessel segmentation. Taking inspiration from Completely Automated Public Turing Test To Tell Computers and Humans Apart, better known as \verb|CAPTCHA| \citep{Ahn2004}, we initially divide the image volume into 2D image patches and we subsequently request the user to identify the patches containing a vessel or part of it. This task is common on websites to differentiate humans from bots, using image \verb|CAPTCHA|s \citep{Ahn2004,elson2007} of natural images. This procedure, which we denote Vessel-\verb|CAPTCHA|, simplifies the annotation process by requiring 2D patch tags indicating the presence of a vessel (a part of it, or multiple vessels) and, thus, avoiding pixel-wise annotations. The user-provided patch tags are subsequently used to synthesize a pixel-wise \toadd{pseudo-labeled} training set in a self-supervised manner \toadd{using a clustering technique}. These two sets are used to train the framework. 

The proposed framework is composed of two networks: a segmentation network and a classification network. \toadd{The segmentation network extracts vessels on a patch basis to tackle the limitations of deep nets in the segmentation of small objects. The final volumetric segmentation is obtained by concatenating the 2D segmented patches. The} classification network is used for two tasks. First, it allows to enlarge the labeled data without the need for further user-provided annotations. Second, it may act as a \toadd{second opinion \citep{Leibig2017,Vrugt2007} that provides a measure of uncertainty} in low quality \toadd{or complex} images.  We \toadd{evaluate the role of} the classification network \toadd{as an expert opinion, where} only the segmentations from patches identified as vessel patches are kept and those classified as non-vessel patches are masked out.

\subsection{Related Work}\label{sec:relatedwork}
\toadd{\subsubsection{3D Brain Vessel Segmentation}
	A} comprehensive collection of methods and techniques for general vascular image segmentation is reviewed in \citep{Lesage2009,Moccia2018}, where they classify different segmentation frameworks according to their characteristic strategies. Classical approaches typically rely on hand-crafted features, with image intensity-derived \citep{Taher2020}, and first \citep{Law2008}, second \citep{Frangi1998,Sato1997} or higher order \citep{Cetin2015} tensor-derived features among the most common. Feature extraction is followed by a vessel extraction scheme, which performs the final segmentation. Notable extraction schemes include deformable models \citep{Klepaczko2016,Zhao2015a}, voting \citep{Zuluaga2014a}, tracking algorithms \citep{Rempfler2015,Robben2016} and statistical approaches \citep{Hassouna2006}. Their main drawbacks are two. First, these methods rely on hand-crafted features that need to be tuned, requiring high expertise to find a good set of parameters. Second, extraction schemes are not fully automatic: many need manual initialization, and the final results typically call for manual correction, specially when images are noisy.

Deep learning techniques have emerged as an alternative to circumvent the difficulties \toadd{of classical approaches. Existing methods have tried to explicitly address the brain vessel tree complexity by designing shallow convolutional neural networks (CNNs) architectures to avoid possible over-fitting \citep{Phellan2017}, or by partitioning the input image volume, while still relying on deeper and more powerful architectures \citep{Kamnitsas2017,Ronneberger2015}. Different partitioning strategies include anatomical regions \citep{Kandil2018}, 2D slices \citep{Ni2020}, 3D  \citep{Phellan2017,tetteh2020} and 2D patches \citep{Livne2019}.} Despite achieving accuracies similar to those of classical approaches, the main limitation towards the broader use of deep learning techniques remains to be the burden linked to pixel-wise data annotation, \toadd{including multi-plane annotations \citep{Phellan2017} or} further pre-processing \citep{Phellan2017,Kandil2018,Livne2019}.

\toadd{Patch-based approaches \citep{Livne2019,tetteh2020} not only aim at reducing the vessel tree's complexity, but they also try} to mitigate the limitations of neural nets in the segmentation of objects occupying small portions of an image. Our work adopts a similar strategy and it builds upon the advantages of 2D patch-based approaches \citep{Livne2019}, thus making vessels cover a significant portion of the patch, while avoiding pixel-wise annotations. 


\toadd{\subsubsection{Limited Supervision for Image Segmentation} D}ifferent strategies have been explored as an alternative to pixel-wise annotation \toadd{\citep{Cheplygina2019,Orting2020,tajbakhsh2020}}, a tedious and time consuming task requiring a high level of expertise. \toadd{These strategies can be roughly classified, according to the type of labels they use, as partial pixel-wise labels, which include incomplete, sparse or noisy pixel-wise labels \citep{tajbakhsh2020}; or as weak labels, which refer to high-level labels and drawing primitives \citep{Cheplygina2019}. 
	
	Partial pixel-wise labels refer to annotations where only a fraction of the pixels of the object of interest are provided \citep{Bai2018,Cicek2016,Liang2019,Ke2020}. These labels can be provided by the user or generated by simpler methods to produce rough segmentation masks. Semi-supervised methods follow different strategies to exploit partially labeled data under the assumption that it is enough to train a segmentation model. \cite{Bai2018} used image registration to propagate user-provided labels over some image slices containing the aorta. \cite{Cicek2016} designed the 3D-Unet to account for sparse and incomplete pixel-wise labels. Other methods resort to iterative stages of refinement \citep{Liang2019,Ke2020}. Although these methods have reported good performances in medical image segmentation \citep{Cheplygina2019}, the complexity of the 3D brain vessel tree makes pixel-wise annotation, even if partial, highly time consuming. As one of our aims is to minimize the annotation effort, our work focuses on the use of weak labels.} 

\toadd{\subsubsection{Weakly Supervised Learning}
	\paragraph{Weak Labels for Medical Image Segmentation} We consider two forms of weak labels for medical image segmentation tasks: image-level labels and drawing primitives.} 
%
\toadd{Image-level labels \citep{Feng2017,Jia2017,Raza2019,Schlegl2015,Xu2019,Zhao2019} assign a tag or rating to an image under the assumption that images contain cluttered scenes with enough information from which a model can learn \citep{qi2017}. In medical tasks, they have been mainly used with 2D images/slices to segment pathologies, i.e. lung nodules \citep{Feng2017}, damaged retinal tissue \citep{Schlegl2015}, brain tumors \citep{izadyyazdanabadi2018} or cancerous tissue \citep{Jia2017,kraus2016,Lerousseau2020,Xu2014,Xu2019}. To a lesser extent they have been used for organ structures segmentation, i.e. the optic disc \citep{Zhao2019}. Despite the good reported performances and the annotation time savings they represent, image tags have not been used for 3D vessel segmentation.
}

\toadd{Drawing primitives include bounding boxes and contouring shapes \citep{Cheplygina2016,Gao2012a,dai2015,Li2018a,Rajchl2017,Wang2018}, scribbles and lines \citep{Can2018,lin2016,matuszewski2018,Wang2015} and clicks \citep{bruggemann2018}. In 3D vessel segmentation,} bounding boxes have been used for aortic segmentation, with the assumption that the aorta is a compact structure, which can be enclosed within a bounding box \citep{Pepe2020}. This assumption does not hold for highly sparse bifurcated trees, as the brain vascular tree, where a 3D bounding box would nearly cover the full brain. Moreover, if an image is analyzed in 2D, the vessel tree appears as a series of disconnected blobs or elongated structures, which challenges the use of 2D contouring shapes. \cite{kozinski2020} address this limitation by using 2D annotations in Maximum Intensity Projections of 3D vascular images. To some extent, these can be considered 2D image scribbles of varying density for the original 3D volume. The framework, however, requires full 2D pixel-wise annotations. Although the scheme significantly reduces the labeling time, more than four hours are needed to generate sufficiently dense 2D annotations that do not compromise performance. \toadd{Finally, clicks are common in classical 3D vessel segmentation approaches \citep{Benmansour2009,Moriconi2019} to provide seed-points, but no works yet integrate them in a weakly supervised learning framework. This may be due to the complexity of the 3D brain vessel tree, where a single click might not carry sufficient information to train a model.} 

\toadd{Our work relies on image tags.}  To cope with the granularity \toadd{and sparse appearance} of vessels, we use 2D patch-level tags, \toadd{in the form of clicks over a grid. A click selects the patches containing at least one vessel or a part of it. We denote this annotation scheme the Vessel-}\verb|CAPTCHA|. 


\toadd{\paragraph{Weakly Supervised Learning with Image Tags}
	Our weakly supervised vessel segmentation framework using image tags can be cast as a multi-instance learning (MIL) problem \citep{dietterich1997,Maron1997,Cheplygina2019}, where a bag corresponds to an image patch and the instances are the image pixels. A bag is considered positive (a vessel patch) if at least one instance within the bag is positive (a vessel pixel). The goal is then to infer the key instances \citep{liu2012}, i.e. the vessel pixels, that activate the bag label.
	
	Standard MIL segmentation approaches, which have been less studied than the classification counterpart \citep{Campanella2019,Hou2016,quellec2012}, follow a multi-stage strategy. In a first stage common to MIL segmentation and classification, they train a model to learn instance-level probabilities of belonging to the positive class. At a second stage, these probabilities are used to obtain pixel-wise labels, which can be considered as the segmentation output \citep{Xu2014,kraus2016} or as pseudo-labels to train a segmentation model in supervised way  \citep{Lerousseau2020,Xu2019}. A main limitation is that the instance-level probabilities are not originally conceived to generate segmentations, but to serve as inputs for bag classification. Therefore, the segmentation results may be poor. Mitigation strategies rely on area constraints \citep{Jia2017,Lerousseau2020}; robust instance selection operations \citep{kraus2016,Xu2019}; post-processing \citep{kraus2016}; or enriched information, such as supplementary instance-level inputs \citep{Shin2019} or image landmarks \citep{Schlegl2015}. However, these strategies often come at the cost of further required user inputs \citep{Jia2017,Schlegl2015,Shin2019}.
	
	Attention-based MIL \citep{Ilse2018}, an alternative to standard MIL, uses attention mechanisms \citep{Niu2021}, such as class activation maps (CAM) \citep{Zhou2016}, under the assumption that the discriminative regions identified by a network correspond to the key instances, i.e. the pixels to segment \citep{Ahn2018,Feng2017,Hong2017,izadyyazdanabadi2018,Ouyang2019,shen2021,Zhao2019}.  Since attention mechanisms focus on the localization of the most discriminative regions, they suffer from the same limitations as standard MIL, which lead to inaccurate segmentation masks. For instance, some works \citep{shen2021} consider the resulting mask as a localization/detection mask and not as a segmentation one. Others have attempted to refine the attention maps through pixel similarity propagation \citep{Ahn2018,Zhao2019}, feature assembling \citep{izadyyazdanabadi2018} and post-processing stages \citep{Krahenbuhl2011}, which all lead to increased model complexity. To avoid the increased complexity, other works propose manual intervention \citep{Feng2017} or the use of some pixel-wise annotated data \citep{Ouyang2019,Zou2021}, leading to more user-required inputs. 
	
	A last set of methods favors the use of simpler techniques to generate an initial pseudo-labeled set that can be then refined using a learning-based approach. \cite{Luo2020} relied on traditional saliency methods along with a quality control step for object detection from videos. \cite{Hou2016} used a mixture of Gaussians in cancer tissue classification. \cite{Lu2021} used a simple threshold to segment tissue regions, which are refined with a CAM to classify cancerous tissue.
	
	While the cerebrovascular tree is a highly complex structure, the typical available dataset size for training a model to segment it is relatively small. Therefore, avoiding high model complexity is critical in 3D brain vessel segmentation \citep{Phellan2017}. Our work favors model simplicity and minimal user interaction. Thus, similarly to \citep{Hou2016,Luo2020,Lu2021}, we use a simpler self-supervised technique, such as the K-means, to generate pixel-wise pseudo-labels. As other weakly supervised approaches \citep{Feng2017,Lerousseau2020,Luo2020,Xu2019}, we use the pseudo-labeled set as input of a supervised training phase that learns to segment the brain vessel tree, without the need for any additional user inputs. 
}

\toadd{\subsubsection{Biomedical Image Classification} O}ur work explores the use of the Unet \citep{Ronneberger2015} and the Pnet  \citep{Wang2019}, two networks originally conceived for medical image \toadd{segmentation}, for the classification tasks of our framework. \toadd{These two networks have been originally designed for image segmentation. Their adaptation to a classification task can be considered as a MIL formulation, where instance-level information, i.e. pixels, are used to predict a bag label, i.e. the patch tag. Similar to} most biomedical classification tasks, \toadd{previous MIL-based biomedical image classification works \citep{Campanella2019,qi2017}} rely on customized versions of VGG-16 \citep{Simonyan2014} and ResNet \citep{He2016}, the most popular architectures for natural image classification. Others \toadd{\citep{Hou2016}} use task-specific architectures adapted from general purpose networks such as end-to-end CNNs. However, no major performance differences are currently found among them \citep{Lundervold2019}.



\subsection{Contributions}\label{sec:contributions}
The contributions of this work are four-fold: 
\begin{enumerate}
	\item we introduce an annotation and segmentation scheme, the Vessel-\verb|CAPTCHA|, to reduce the labeling burden of 3D brain vascular images, consisting of two phases: a first phase where the user provides tags at the 2D image patch-level, and a second stage where pixel-wise \toadd{pseudo-}labels are \toadd{obtained}, in a self-supervised fashion, using only the user-provided patch tags as input.
	\item We propose a weakly supervised learning framework on 2D image patches to achieve 3D brain vessel segmentation. To circumvent the problems faced by deep neural networks when segmenting small objects, the framework uses a 2D patch-based segmentation network trained with 2D pixel-wise \toadd{pseudo-}labeled patches synthesize by the Vessel-\verb|CAPTCHA| annotation scheme \toadd{using the weak user-provided patch tags as input}. 
	\item We investigate the use of network architectures specifically designed for medical imaging tasks to classify 2D image patches (vessel vs. non-vessel). The classifier networks are used to \toadd{pseudo}-label a potential training set without further user effort, and it may act as a \toadd{second opinion} for segmentation masks \toadd{obtained from low quality images}. 
	\item  Using two different image modalities, we demonstrate that the proposed framework achieves state-of-the-art performance for 3D brain vessel segmentation, while significantly reducing the annotation burden by \toadd{$\sim$77\%} compared to the annotation time required in other deep learning-based methods.
\end{enumerate}

To foster reproducibility and encourage other researchers to build upon our results, the source code of our framework made publicly available on a Github repository\footnote{\url{https://github.com/robustml-eurecom/Vessel-Captcha}}. 

\section{Method}\label{sec:method}
\begin{figure*}[t]
	\centerline{\includegraphics[width=\textwidth]{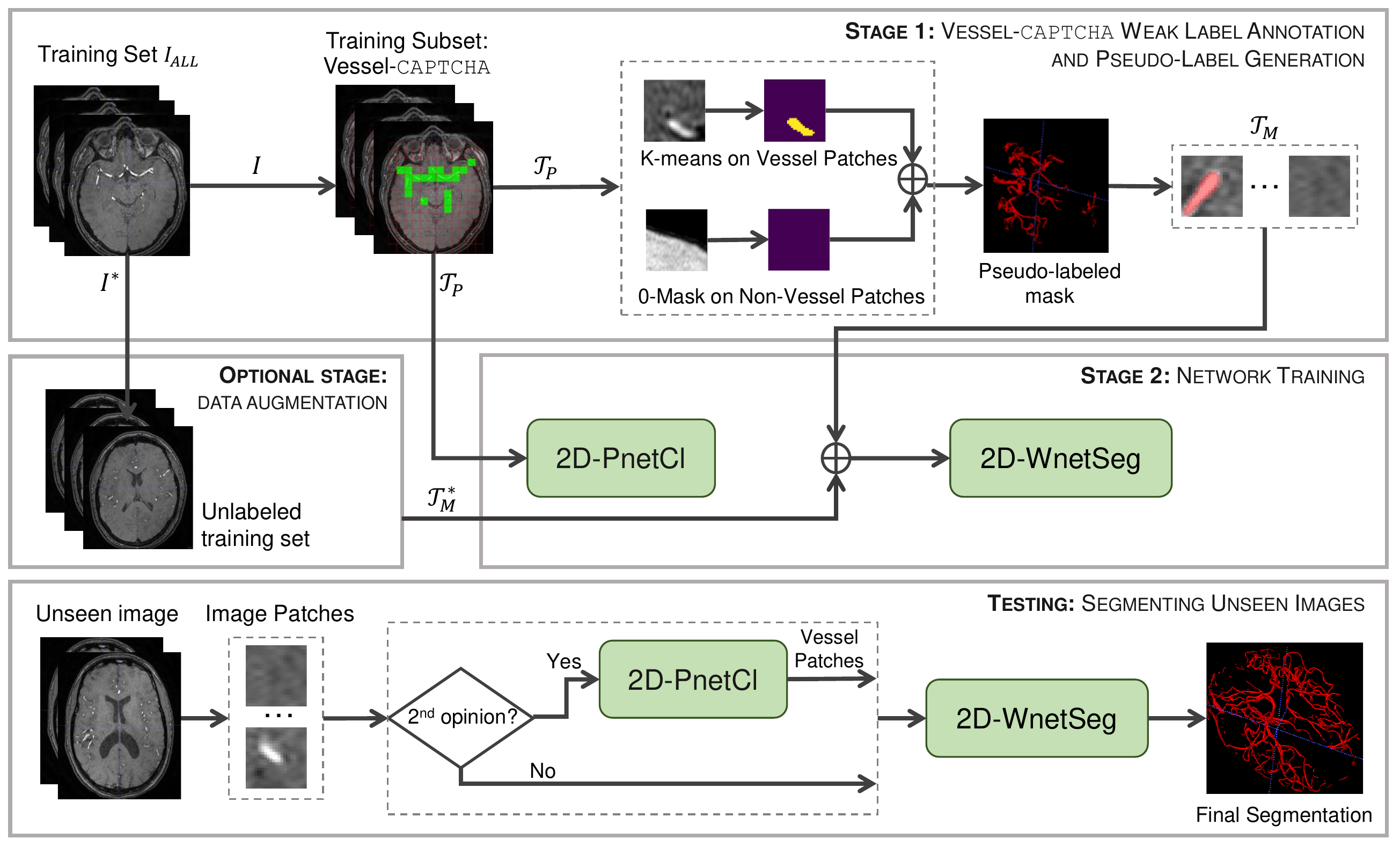}}
	\caption{The Vessel-CAPTCHA framework. 
		\toadd{
			At Stage 1, an image grid with patch size 32$\times$32 covering the brain tissue is presented to the user for annotation. The user selects the patches which contain at least one vessel or a part of it. The process, which we denote the Vessel-CAPTCHA annotation scheme, is done for every axial slice in an image volume. This weakly annotated set $\mathcal{T}_P$ is used to synthetize pixel-wise pseudo-labels for every patch using the K-means algorithm. The resulting pseudo-labeled set is denoted $\mathcal{T}_M$. At stage 2, $\mathcal{T}_P$ is used to train a classification network (2D-PnetCl) and  $\mathcal{T}_M$ is used to train a segmentation network (2D-WnetSeg). In the segmentation network training, it is possible to enlarge the set of pseudo-labeled data through an optional data augmentation step.
			For an unseen image, the final volumetric segmentation is obtained by concatenating the 2D segmentations obtained from 2D-WnetSeg. Optionally, the classification network can be used as a second opinion to refine the segmentation results. In that case, only 2D segmentations from patches classified as vessel ones are considered in the final volume segmentation.}} 
	\label{fig:pipeline}
\end{figure*}
The proposed Vessel-\verb|CAPTCHA| framework algorithm for 3D vessel segmentation is depicted in Fig.~\ref{fig:pipeline}. In the following, we introduce the Vessel-\verb|CAPTCHA| annotation scheme and we describe how pixel-wise \toadd{pseudo-}labels are synthesized from the user-provided \toadd{weak} patch labels in a self-supervised way (Sec.~\ref{sec:annotation_phase}). In Sec.~\ref{sec:training_phase}, we present the two networks conforming the proposed framework: a classifier network and a segmentation network. Sec.~\ref{sec:augment} explains how the classifier network can be used to enlarge the set of weak pixel-wise annotations, allowing to have a larger set to train 2D-WnetSeg. Finally, Sec.~\ref{sec:testing_phase} briefly explains how to segment unseen images using the proposed framework. 

\subsection{The Vessel-CAPTCHA Annotation Scheme}\label{sec:annotation_phase}
We consider a dataset $\mathcal{I}$ of training images. Given an image $\mathbf{I}\in\mathcal{I}$ of size $H \times W \times S$, for each slice $X_s, \, s \in [1,\ldots, S]$, we consider a partition in $P_s$ non-overlapping patches: $\mathcal{X}_s = \{ \hat{X}_k\}_{k=1}^{P_s}$. Each patch is here considered as a function $\hat{X}_k:D_k\rightarrow{}\mathbb{R}$, where $D_k$ is a subset of the slice domain $D_k\subset [1,H] \times [1,W]$. 

User annotations on a given patch $\hat{X}_k$ are defined through a function $U_k:D_k\rightarrow{}\{0,1\}$, assigning a binary label to each coordinate $(i,j)\in D_k$. The set of annotations for a given patch is summarized by an indicator function $f:U_k\rightarrow{}\{0,1\}$ which takes value 1 if at least one pixel in the patch was labeled with 1: 
\begin{equation}
	f(U_k) = 1 \Longleftrightarrow \exists (i,j)\in D_k\, s.t.\, U_k(i,j)=1.  
\end{equation}
\begin{figure}[!t]
	\centerline{\includegraphics[width=\columnwidth]{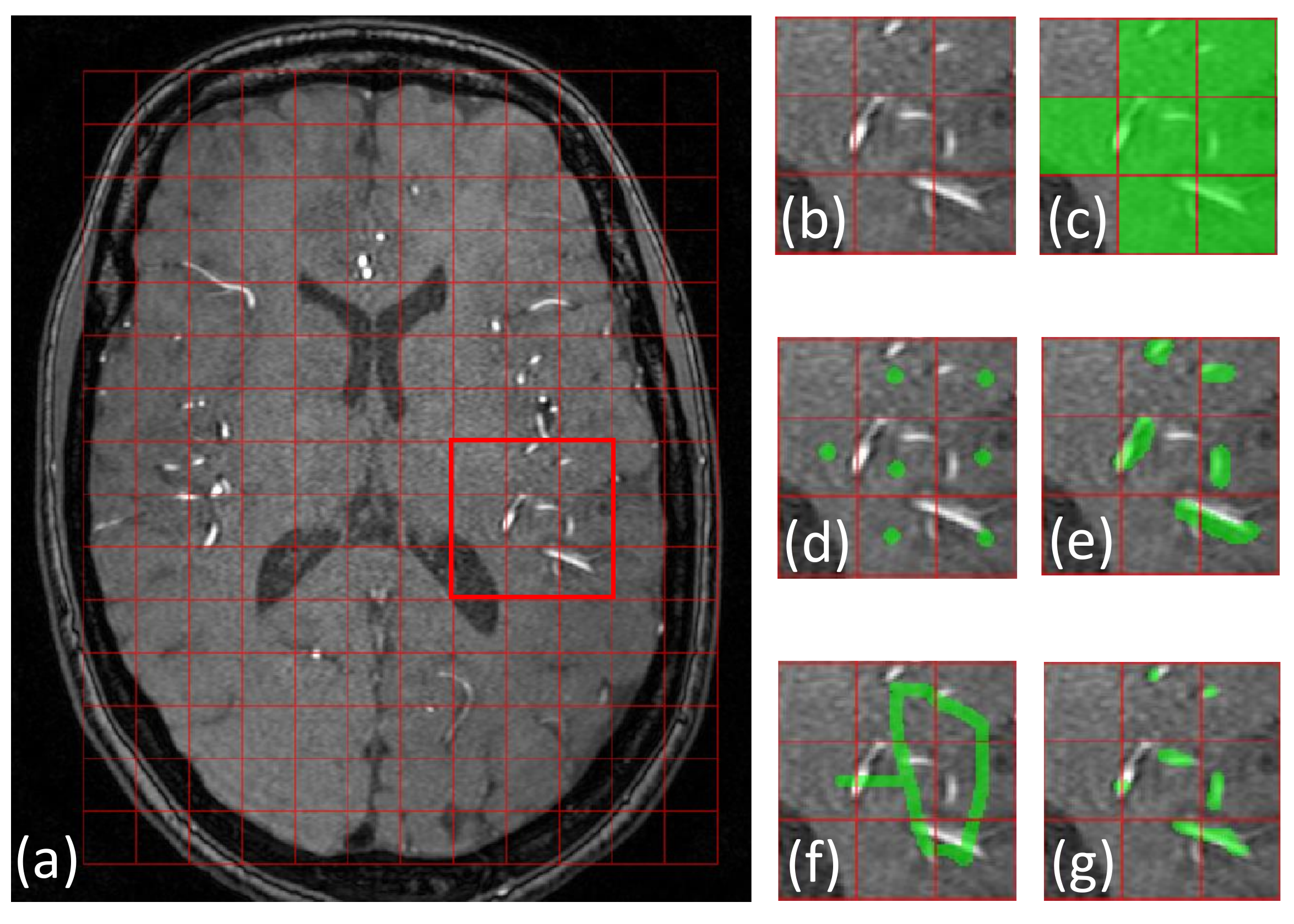}}
	\caption{Example of equivalent CAPTCHA annotations. (a) Image slice $\mathcal{X}_s$ with patch grid, (b) zoomed region corresponding to the highlighted red box in (a), (c) resulting $\mathcal{T}_P$ obtained through equivalent annotations (d-g).} 
	\label{fig:captchas}
\end{figure}
Fig.~\ref{fig:captchas} illustrates examples of equivalent annotations.
The set of indicators for the slice $X_s$ is denoted by $\mathcal{Y}_s = \{f(U_k)\}_{k=1}^{P_s}$. 
The training set of patch-level labels for the image $\mathbf{I}$ is defined by the set:
$\mathcal{T}_P^\mathbf{I} = \{\mathcal{X}_s , \mathcal{Y}_s\}_{s=1}^S$. This set is therefore composed by patches and associated indicators/tags of the presence of a vessel according to the user's annotation.
Based on the training set $\mathcal{T}_P^\mathbf{I}$, we estimate approximated vessel masks via a model fitting procedure. For every patch we define a function $M_k:D_k\rightarrow\{0,1\}$, which assigns to each pixel's coordinate a label according to the following scheme:

\begin{equation}
	M_k(i,j) = \begin{cases}
		0 & \text{if $f(U_k) = 0$},\\
		KM( \hat{X}_k(i,j)) &\text{otherwise},
	\end{cases}
\end{equation}
where $KM$ is a K-means predictor trained on the intensity values of the patch $\{\hat{X}_k(i,j), \,\, (i,j)\in D_k\}$. By specifying $K=2$ clusters we therefore obtain a rough estimate of the low-high intensity partitioning of the patch. The ensemble of estimated partitions across patches is denoted as $\mathcal{M}_s = \{M_k\}_{k=1}^{P_s}$, and we define the pixel-wise labeled training set for the image $\mathbf{I}$ as $\mathcal{T}_M^\mathbf{I} = \{\mathcal{X}_s , \mathcal{M}_s\}_{s=1}^S$.

Finally, patch- and pixel-level training sets across the image dataset are denoted by 
\begin{equation}\label{eq:training_patch}
	\mathcal{T}_P = \{\mathcal{T}_P^\mathbf{I}\}_{\mathbf{I}\in\mathcal{I}},
\end{equation}
and 
\begin{equation}\label{eq:training_mask}
	\mathcal{T}_M = \{\mathcal{T}_M^\mathbf{I}\}_{\mathbf{I}\in\mathcal{I}},
\end{equation}
respectively.

\subsection{Image Segmentation and Patch Classification Networks}\label{sec:training_phase}

\subsubsection{Segmentation Network} 
\begin{figure}[t]
	\centerline{\includegraphics[width=\columnwidth]{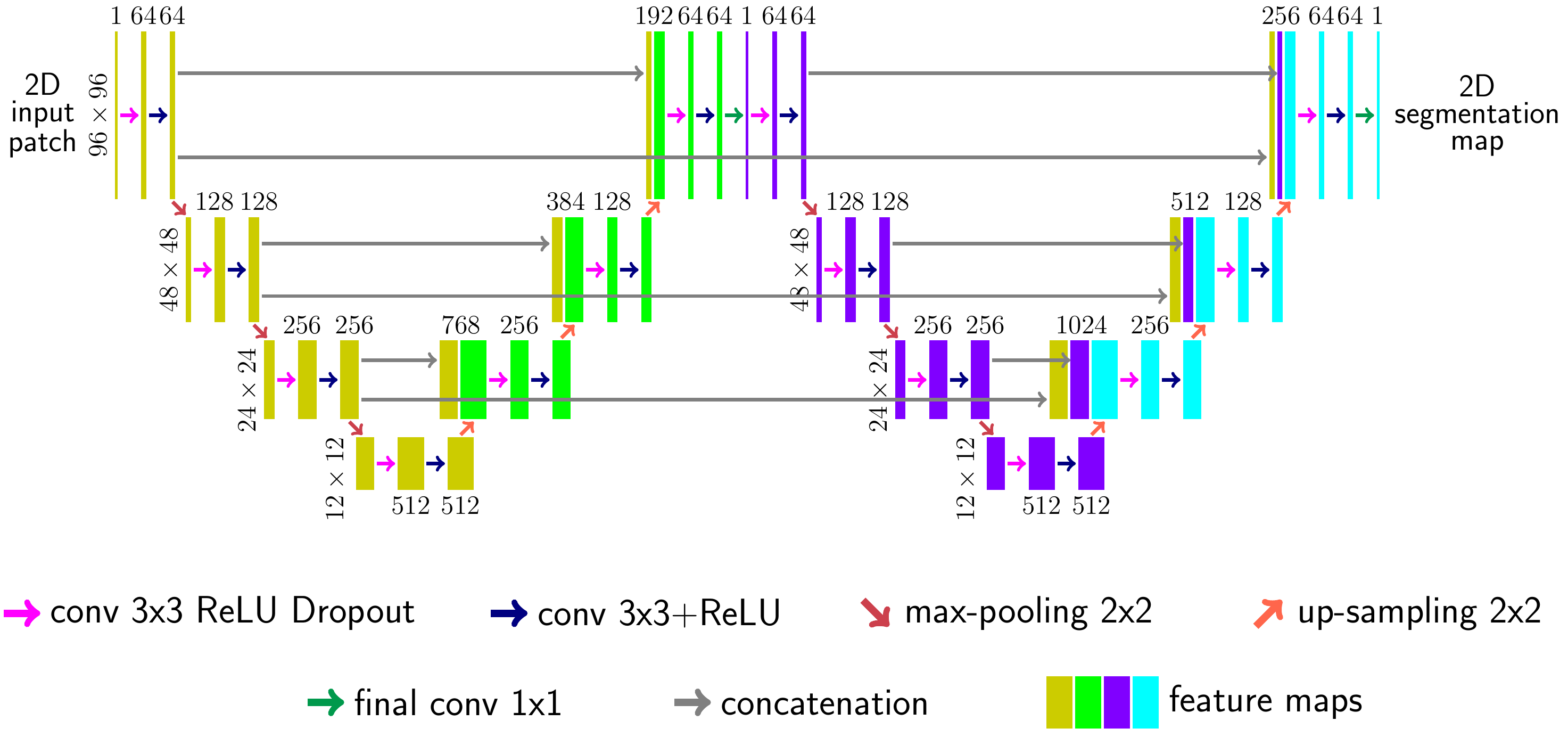}}
	\caption{Illustration of the 2D-WnetSeg architecture.} 
	\label{fig:wnet}
\end{figure}

\toadd{The segmentation network learns from the input training set $\mathcal{T}_M$ how to segment 2D image patches using} the Dice \toadd{similarity} coefficient, as proposed by \cite{Milletari2016}, which is specifically tailored for segmentation tasks in medical images. \toadd{The segmented 2D patches are concatenated to reconstruct the original segmented 3D image volume. For this task,} we use a segmentation network connecting two 2D-Unets in cascade \citep{dias2019}. We denote it 2D-WnetSeg (Fig.~\ref{fig:wnet}). The network is trained on $\mathcal{T}_M$, the set of 2D image patches with pixel-wise \toadd{pseudo-}labels to tackle the neural networks limitations in the segmentation of objects with a small object-to-image ratio. 

The human cerebrovascular system has an intricate shape with large and smaller blood vessels which mainly differ in the spatial scale, but which share similar shapes. The selected self-supervised method, the K-means, favors the over-segmentation of larger vessels. Thanks to a set of max pooling layers, the first 2D-Unet allows to learn spatial scaling features from the input training data. Thus, it can recover rough-mask labels from smaller vessels not initially extracted by K-means. This means that the first Unet acts as a refinement module to correct the initial masks by inferring missing vessels based on the structural redundancy of the cerebrovascular tree. The second Unet, which has a similar architecture as the first one, receives as input the output of the first Unet with the recovered labels from small vessels. As a result, the 2D-WnetSeg is able to learn vessels even with a \toadd{pseudo-}labeled training set with imperfect labels or noise. 

The smaller vessels in the brain vessel tree may disappear in very deep networks due to the subsampling layers. To tackle this, the 2D-WnetSeg has 14 blocks with convolutional layers structured into 4 levels. In this, it differs from previously proposed cascaded networks \citep{dias2019} or the Unet-based vessel segmentation from \citep{Livne2019}. This also contributes to reduce the number of trainable parameters. Specifically, the number of trainable parameters in \citep{Livne2019} is about $3.1\mathrm{e}7$, whereas the WnetSeg has only about $1.6\mathrm{e}7$ parameters. 

In our architecture, the first 7 blocks form the first Unet and the second 7 blocks belong to the second one. Each block consists of 2 convolutional layers with kernel size $3\times3$ pixels, each followed by a rectified linear unit (ReLU). They are both added to the padding to ensure that the output has the same shape as the input. A drop-out layer is applied between them. As the input proceeds through different levels along the contracting path, its resolution is reduced by half. This is performed through a $2\times2$ max-pooling operation with stride 2 on 3 levels except for the bottom level. We double the number of feature channels at each level of the contracting path. The right portion of a half-network (Unet), i.e. the expansive path, consists of blocks with concatenation and up-sampling for each level to extract low-features and it expands the spatial support of the lower resolution feature maps to assemble the necessary information and recover the original input size. Finally, we employ skip-connections from the shallow layers to deeper layers between the two 2D-Unets, at the same levels, to ease the training of the network.

\subsubsection{Networks for vessel vs. non-vessel patch classification}
The classification network is trained on $\mathcal{T}_P$ to discriminate between vessel and non-vessel patches in unseen data. This discrimination serves two purposes: 1) to synthesize patch tags without the need of user interventions and 2) to act as a \toadd{second opinion} for segmentations. \toadd{In the latter case, the segmentation network serves as a first expert predicting pixel-wise labels, whereas the classifier network provides a concept on a per-patch basis. This can be considered an ensemble approach to uncertainty \citep{Vrugt2007}, where a disagreement among the two networks/opinions indicates uncertainty on the predictions of a given patch.}

Most works in the literature rely on customized versions of VGG-16 \citep{Simonyan2014} and ResNet \citep{He2016}, the most popular architectures for natural image classification, or on task-specific architectures adapted from general purpose networks \citep{Chen2016,Setio2016}. In this work, we investigate the use of networks specifically designed for medical imaging applications for our classification task: the Unet \citep{Ronneberger2015} and the Pnet \citep{Wang2019}. As these two networks have been designed for image segmentation, we hereby describe how they have been modified to achieve classification. 

We denote the modified 2D Pnet architecture \citep{Wang2019} 2D-PnetCl. It consists of 7 convolution layers, 2 dropout layers, and a sigmoid layer. The first 5 convolution layers are concatenated. Each convolutional layer contains 64 filters with 3$\times$3 \toadd{pixel}s receptive fields in a 1 \toadd{pixel} stride sliding with different dilation factors. The dilations are 1, 2, 4, 8 and 16, respectively. The last two convolutional layers are the $1\times1$ convolutions, the output feature map is flattened and fed to a fully connected layer for interpretation with 128 hidden units and the final prediction layer uses a sigmoid function with one unit to classify patches with and without vessels. The adapted 2D-Unet architecture, denoted 2D-UnetCl, uses the network from \citep{Livne2019} as a starting point. Similarly to the 2D-PnetCl, the output feature map is flattened and fed to a fully connected layer for interpretation with 128 hidden units and a final prediction layer with one unit to classify patches with and without vessels.

\subsection{Data Augmentation for Segmentation Network Training}\label{sec:augment}
\begin{figure}
	\centerline{\includegraphics[width=\columnwidth]{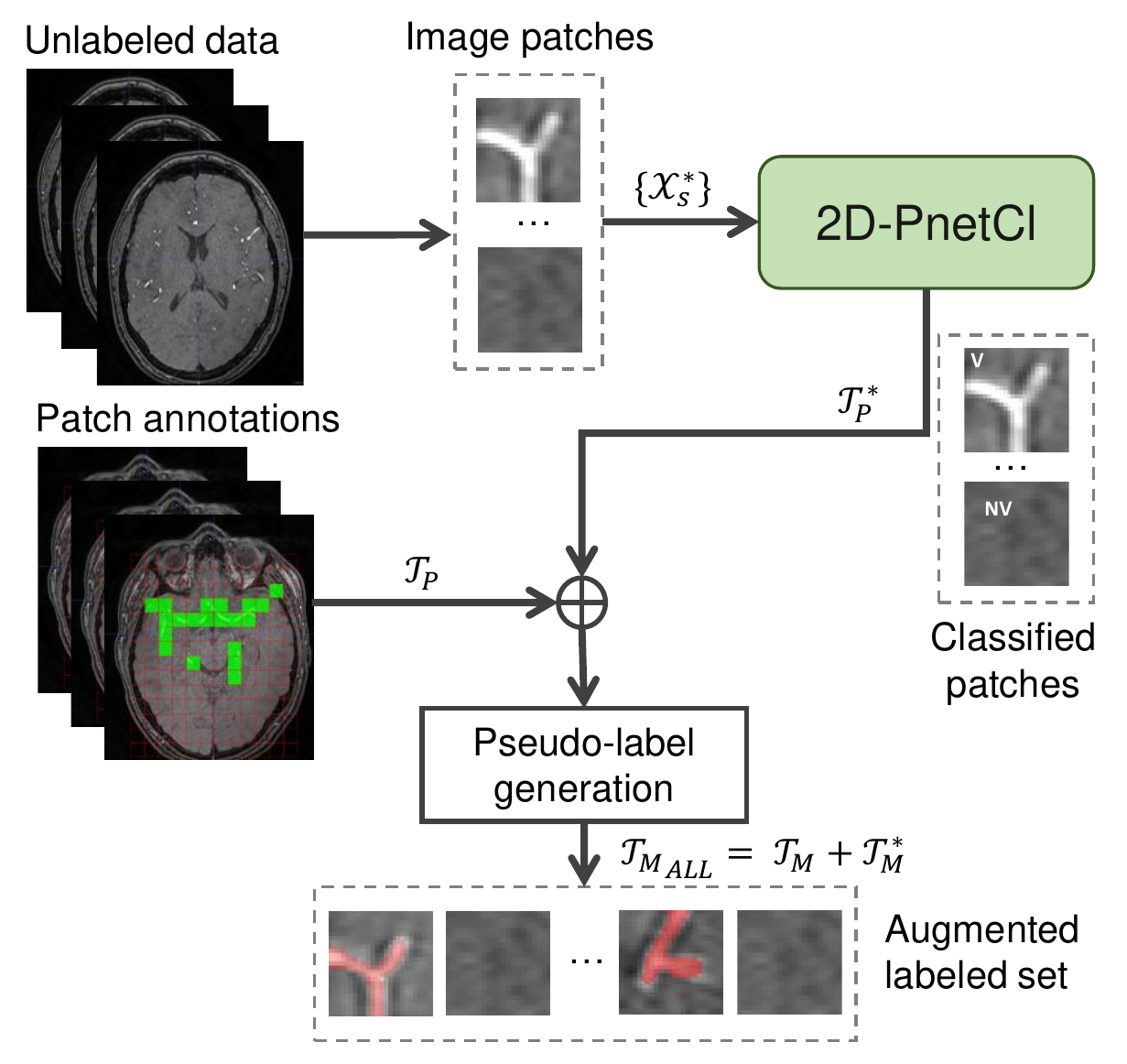}}
	\caption{Data Augmentation procedure. The trained classifier is used as the starting point to enlarge the initial pixel-wise labeled training set $\mathcal{T}_M$ without requiring further user inputs. The resulting training set $\mathcal{T}_{M_{ALL}}$ is a combination of both the \toadd{pseudo-}labels and those obtained via the Vessel-CAPTCHA annotation.} 
	\label{fig:trainingplus}
\end{figure}

The set $\mathcal{T}_M$ consisting of \toadd{pseudo-labels} is used to train the 2D-WnetSeg. To augment its size without increasing the annotation burden, we make use of the classification network to generate a larger set with \toadd{pixel-wise pseudo-labels}. The procedure is depicted in Fig.~\ref{fig:trainingplus}. 

Assuming that there is an initial set of unlabeled images ${I}^*$ that can be used for training, we consider the joint image dataset of labeled and unlabeled images $\mathcal{I}_{ALL}=\mathcal{I}\bigcup\mathcal{I}^*$. The subset $\mathcal{I}$ of these images is used to generate Vessel-\verb|CAPTCHA|s, which are presented to the user for annotation. This results in the training set $\mathcal{T}_P$ (Eq.~\ref{eq:training_patch}), which is used to both train the classification network and to synthesize the pixel-wise \toadd{pseudo-}labeled set $\mathcal{T}_M$ (Eq.~\ref{eq:training_mask}). 

Using the trained classification network, a set of patches $\{\mathcal{X}^{*}_{s}\}$ is obtained in the remaining set of images $\mathcal{I}^{*}$. Rather than presenting another Vessel-\verb|CAPTCHA| to the user for annotation, the $\{\mathcal{X}^{*}_{s}\}$ are inputted to the classification network to estimate patch labels $\{\mathcal{Y}^{*}_{s}\}$. The paired set of patches and estimated labels conform a new set $\mathcal{T}^{*}_P=\{\mathcal{T}_P^{\mathbf{I}}\}_{\mathbf{I}\in\mathcal{I}^*}$.

The set $\mathcal{T}^{*}_P$ is used to synthesize pixel-wise \toadd{pseudo-}label masks ${\mathcal{M}^*}$ following the same procedure applied to $\mathcal{T}_P$ (Sec.~\ref{sec:annotation_phase}). 
This leads to a new \toadd{pseudo-}labeled set $\mathcal{T}_{M}^{*}$.
The extended set of pixel-wise \toadd{pseudo-}labels is formed by the union of the two sets $\mathcal{T}_{M_{ALL}}=\mathcal{T}_{M} \bigcup \mathcal{T}_{M}^{*}$, and  is subsequently used to train the 2D-WnetSeg architecture.

\subsection{Inference Phase}\label{sec:testing_phase}
Unseen 3D images are segmented by extracting 2D image patches that are then segmented by the 2D-WnetSeg and concatenated to build back the original volume (Fig.~\ref{fig:pipeline}). In \toadd{low quality or} noisy images, the resulting segmentation can often present a large set of pixels erroneously segmented as vessels. 
To avoid this problem, the trained classifier network \toadd{may act as an expert providing a second opinion to the results from the segmentation network}. In such case, only those patches which have been classified as vessels are taken into account to reconstruct the final volume. All the pixels of the remaining patches are set to zero. 

\subsection{Implementation Details}
We used the Keras library to implement 2D-PnetCl, 2D-UnetCl and 2D-WnetSeg. The networks were trained on a GPU workstation with 4-core Intel(R) Xeon(R) CPU @ 2.30GHz, a NVIDIA Tesla P100-PCIE-16GB, and 25GB memory. For both 2D-UnetCl and 2D-PnetCl we optimized the binary cross-entropy loss function with a minibatch stochastic gradient descent and a conservative learning rate of 0.01 and momentum of 0.9. The weights of the 2D-WnetSet were optimized using an Adam optimizer with learning rate $lr=1\mathrm{e}{-4}$, $\beta_{1}=0.9$, and $\beta_{2}=0.999$. All networks were trained from scratch using mini-batches of 64 patches. All input patches were normalized by the mean and standard deviation of the whole training data. 
A dropout of 0.5 for 2D-PnetCl and 2D-UnetCl, and of 0.1 for 2D-WnetSeg was added to prevent overfitting during the training. The image input sizes of 2D-PnetCl and 2D-WnetSeg were 32$\times$32 and 96$\times$96, respectively. We implemented a zero-padding technique to preserve output size as input size at each convolution layer in both networks. Therefore, the feature map size at each level in the 2D-PnetCl is 32$\times$32. 

\toadd{
	\section{Experimental Setup}
	\label{sec:experiments}
	In this section, we describe the experimental setup. First, we present the datasets used in our experiments (\ref{sec:data}) and the baselines used for comparison (Sec. \ref{sec:benchmarks}). Then, we describe the training setup (\ref{sec:setup}). Finally, we present the performance evaluation metrics used in our experiments (Sec. \ref{sec:metrics})}.

\subsection{Data}\label{sec:data} 
\toadd{Three different types of data were used in this study:  synthetic, Time-of-Flight (TOF) angiography and Susceptibility-Weighted Images (SWI). The latter two correspond to two magnetic resonance imaging (MRI) sequences} commonly used to image and assess the cerebrovascular tree \citep{Radbruch2013}, \toadd{although} blood vessels present different appearances in each modality. In TOF, vessels are hyper-intense structures, whereas they are hypo-intense in SWI. 
\toadd{Table~\ref{tab:dataset_summary} summarizes the main properties of each data type and the datasets used.}

\begin{table*}[t]
	\centering
	\caption{\toadd{Main properties of data used and training and validation test sizes per data type}}
	\setlength{\tabcolsep}{5pt}
	\begin{tabular}{|l|r|r|r|}
		\hline
		& \multicolumn{1}{|c|}{\toadd{Synthetic}} & \multicolumn{1}{|c|}{\toadd{TOF}} 
		& \multicolumn{1}{|c|}{\toadd{SWI}} \\
		\hline
		\toadd{Dataset size} & \toadd{136} & \toadd{100} & \toadd{33} \\
		\toadd{Volume dimensions} & \toadd{$325\times304\times600$} & \toadd{$560\times560\times117$ (Set 1)}& \toadd{$480\times480\times288$}\\
		& & \toadd{$576\times768\times232$ (Set 2)}& \\
		\toadd{Voxel spacing}& \toadd{$1\times1\times1$ mm$^{3}$}& \toadd{$1\times1\times1$ mm$^{3}$ (Set 1)} &  \toadd{$1\times1\times1$ mm$^{3}$}\\
		& & \toadd{$0.3\times0.3\times0.6$ mm$^{3}$(Set 2)} & \\
		\toadd{$|\mathcal{T}_P|$ (patch size $32\times32$)} & \toadd{7.18M}& \toadd{770K} & \toadd{30.6K}\\
		\toadd{$|\mathcal{T}_M|$ (patch size $96\times96$)} & \toadd{1.04M}& \toadd{110K} & \toadd{10.2K}\\
		\hline
	\end{tabular}
	\label{tab:dataset_summary}
\end{table*}

\toadd{
	\paragraph{Synthetic Data} We use the synthetic data generated and made public by \cite{tetteh2020}\footnote{ \url{https://github.com/giesekow/deepvesselnet/wiki/Datasets}}. The dataset consists of 136 volumes of size $325\times304\times600$ with corresponding labels for vessel segmentation, which were generated following the method proposed in \citep{Schneider2012}. The vessel labels occupy 2.1\% of total intensities, highlighting the problem of vessels being relatively small objects within a large image volume.
	
	\paragraph{TOF Data} We use 100 TOF scans coming from two different sources.} Forty-two TOF subject scans, from retrospective studies previously conducted at the UCL Queen Square Institute of Neurology, were available with volume dimensions $560\times560\times117$ and isotropic voxel size $1\times1\times1$ mm$^{3}$\toadd{(Set 1). The remaining 68 scans were obtained from the OASIS-3 database \citep{LaMontagne2019} with volume dimensions $576\times768\times232$ and voxel size $0.3\times0.3\times0.6$ mm$^{3}$ (Set 2).}

\toadd{\paragraph{SWI Data} We }use 33 different subject scans with image dimensions $480\times480\times288$ and isotropic image resolution $1\times1\times1$mm$^{3}$, from retrospective studies previously conducted at the UCL Queen Square Institute of Neurology, Queen Square MS Centre, University College London. Due to poor image quality, three SWI scans were discarded for the experiments.

\toadd{
	\subsection{Baselines}\label{sec:benchmarks} 
	We compare our segmentation framework with several alternatives, including state-of-the art deep learning-based vessel segmentation \citep{Livne2019,tetteh2020} and classical approaches \citep{Frangi1998,Sato1997,Zuluaga2014a}, and weakly supervised learning frameworks \citep{Ahn2018,Lerousseau2020}.  Specifically, we evaluate:}
\begin{enumerate}
	\item \toadd{\textbf{Classical 3D Vessel Segmentation Methods:}} We consider three classical non-learning based approaches, \toadd{which use the 3D image volume as input}. These are: the Frangi filter \citep{Frangi1998} (\textbf{Frangi}) and the Sato filter \citep{Sato1997} (\textbf{Sato}), two references for vessel segmentation, and a tensor voting framework for 3D brain vessel segmentation \citep{Zuluaga2014a} (\textbf{TV}).
	\item \toadd{\textbf{Deep Leaning-based 3D Vessel Segmentation Methods:} We consider} the deep learning-based brain vessel segmentation framework from \cite{Livne2019} (\textbf{Vessel 2D-Unet}), which relies on the 2D-Unet \citep{Ronneberger2015} as backbone architecture, and uses 2D patches as input; \toadd{and \textbf{DeepVesselNet}, the framework from \cite{tetteh2020}, which uses the 3D image volume as input, but operates on 3D patches using a fully convolutional architecture to extract the 3D vessel tree.}
	\item \toadd{\textbf{Weakly Supervised Methods:} We compare our weakly supervised strategy with one standard MIL and a CAM-based approach. Concretely, we use a MIL framework for whole slice (\textbf{WS-MIL}) histopathology segmentation \citep{Lerousseau2020} and the CAM-based approach proposed by \cite{Ahn2018} for natural image segmentation (\textbf{AffinityNet}). Both methods work with 2D image patches with size 32$\times$32 and 96$\times$96, respectively.} 
	\item \toadd{\textbf{Other Limited Supervision Strategies:} We consider two semi-supervised strategies using partial labeling. The \textbf{3D-Unet} \citep{Cicek2016}, which has been designed to account for sparse annotations of a 3D image volume, and a \textbf{Pseudo-labeling} strategy, where we use rough masks as labels. The label masks are generated with the Sato filter \citep{Sato1997} and they are used to train a 2D-Unet network with 2D image slices.}
\end{enumerate}

We compare the classification networks, 2D-PnetCl and 2D-UnetCl, with two baselines, \textbf{VGG-16} \citep{Simonyan2014} and \textbf{ResNet} \citep{He2016}, as they are among the most common networks for classification \toadd{\citep{Litjens2017}}. Table~\ref{tab:others} summarizes the hyperparameter setup for every baseline network.
%


\begin{table}[t]
	\centering
	\caption{Hyper-parameter setup for baseline networks}
	\small
	\setlength{\tabcolsep}{5pt}
	\begin{tabular}{|p{57pt}|p{170pt}|} 
		\hline
		
		Network & Hyper-parameters\\
		\hline
		Vessel 2D-Unet & batch size: 64, lr: 1$\mathrm{e}$-4, dropout: 0.0\\
		\toadd{DeepVesselNet} & \toadd{batch size: 10, lr: 1$\mathrm{e}$-3, decay: 0.99, cube size: 64}\\
		\toadd{WS-MIL} & \toadd{batch size: 100, lr: 1$\mathrm{e}$-4, decay: 10$\mathrm{e}$-5, $c_0$=$c_1$=1,  $\alpha=[1\mathrm{e}$-2$,\ldots,0.1],$  $\beta=0.00$}\\
		\toadd{AffinityNet} & \toadd{batch size: 16, lr: 1$\mathrm{e}$-1} \\
		3D-Unet  & lr: 1$\mathrm{e}$-4, reduced by 0.5 every 10 epochs. Stopped at 50 epochs if no improvements in the validation error\\
		\toadd{VGG-16} &  \toadd{batch size: 64, lr: 1$\mathrm{e}$-4}\\
		\toadd{ResNet} &  \toadd{batch size: 64, lr: 1$\mathrm{e}$-3}
		\\
		
		\hline
	\end{tabular}
	\label{tab:others}
\end{table}

\toadd{
	\subsection{Setup}\label{sec:setup} }
\toadd{\paragraph{Pre-processing and Annotation} 
	We used the available ground truth from the synthetic images to generate Vessel-}\verb|CAPTCHA| \toadd{annotations. Since the in-plane dimensions of the images are not a multiple of the patch size (Table~\ref{tab:dataset_summary}), we overlap the last two rows/columns of patches. 
	
	Both TOF and SWI} were skull-stripped using a standard tool and we generated the Vessel-\verb|CAPTCHA| annotation grid only over the brain tissue (Fig.\ref{fig:captchas}). \toadd{Where the minimum-sized rectangle mask covering the brain tissue was not a multiple of the patch size in a given dimension, we dilated the mask in that dimension until the condition was met and generate the annotation grid. If the minimum-sized rectangle mask touched the image slice borders and the in-plane dimensions of the images were not a multiple of the patch size, we generated the annotation grid by overlapping the last two rows or columns of patches. Three users} annotated the images using the Vessel-\verb|CAPTCHA| annotation scheme\toadd{: a trainee, an experienced rater and a neurologist.} In addition to this, TOF data was pixel-wise annotated. Finally, no pixel-wise labels were obtained for SWI, since it is difficult to obtain a sufficiently robust ground truth. \toadd{All annotation times were recorded.}

For the Vessel 2D-Unet, further data pre-processing for synthetic and TOF data was performed as described in \citep{Livne2019}. \toadd{All datasets where normalized (within modality). For TOF, where two different sources were used, we follow the intensity and spacing normalization strategy from \citep{Full2021}.}

\toadd{\paragraph{Training Setup} Table~\ref{tab:dataset_summary} displays the number of available 2D patches for training and validation per dataset.} For every dataset, we performed data splitting at the image volume level, using \toadd{ a split ratio 70/10/20\%} for training, validation and testing, respectively. The training sets were augmented through the use of different random rotations, flips and shears at every epoch for every 2D patch.  Models are chosen based on the best performance in the validation set. 

Two different rules are used to synthesize \toadd{pseudo-labels for} the annotated training set $\mathcal{T}_M$ with the K-means algorithm. In \toadd{synthetic data and} TOF, vessels are associated to the cluster with the highest mean value, whereas the vessel class is associated to the cluster with the lowest mean value in SWI. The training sets, $\mathcal{T}_P$ and $\mathcal{T}_M$, are used to separately train a classification and a segmentation network per modality.

\toadd{
	\subsection{Evaluation Metrics}\label{sec:metrics} 
	\paragraph{Vessel Segmentation} We estimate the Dice Similarity Coefficient (DSC), the Hausdorff Distance (HD), the 95\% Hausdorff Distance (95HD) and the mean surface distance error ($\mu$D)} between the segmentation and the annotated ground truth to quantitatively assess the segmentation accuracy \toadd{in TOF and the synthetic dataset. We measure HD, 95HD and $\mu$D in voxels}.

\toadd{In SWI}, the segmentations are assessed qualitatively. Based on a visual inspection \toadd{by two raters (an expert rater and a neurologist)}, the segmented images are classified as good \toadd{(3)}, average \toadd{(2)} or low quality \toadd{(1)}.
A segmented image is considered good, if it segments the large and medium vessels, and avoids the segmentation of noisy regions, with an elongated appearance similar to a vessel, and sulci. It might miss some small vessels. A segmented image is considered of average quality if it segments large and medium vessels, it misses small ones, it may segment noisy areas in a small proportion (less than 50\%), specially in the anterior part of the brain, and often segments sulci. All other cases are considered as low quality ones. \toadd{We use the Cohen's Kappa coefficient ($\kappa$) to measure the level of agreement among raters.} 

\toadd{\paragraph{Patch Classification} We measured }precision (P), recall (R) and 
the F-score (F$_{1}$), using a vessel patch as the positive class to assess the quality of the classification results obtained by \toadd{the classifier networks}.

%


\toadd{
	\section{Experiments and Results}
	\label{sec:results}
	We assess the performance of the Vessel-}\verb|CAPTCHA| \toadd{in terms of vessel segmentation accuracy and required annotation time (Sec.~\ref{subsec:vessel}). In Section~\ref{subsec:weak}, we compare our weak learning strategy with other limited supervision techniques. Section~\ref{subsec:classification} studies the proposed classification networks and their performance as a data augmentation strategy.} Next, we perform an ablation study to understand how the different components of the framework contribute to performance \toadd{ (Sec.~\ref{subsec:ablation}) and we present a brief summary of all the obtained results in Section~\ref{subsec:summary}}. 

\toadd{
	\subsection{3D Brain Vessel Segmentation Performance}\label{subsec:vessel}
	We evaluate the performance of the Vessel-}\verb|CAPTCHA| \toadd{framework in terms of segmentation accuracy and required annotation time using all available datasets.} We compare it against the 3D brain vessel segmentation, i.e. the deep learning vessel segmentation frameworks and the classical techniques. 

\begin{figure}[t]
	\centerline{
		\includegraphics[width=0.8\columnwidth]{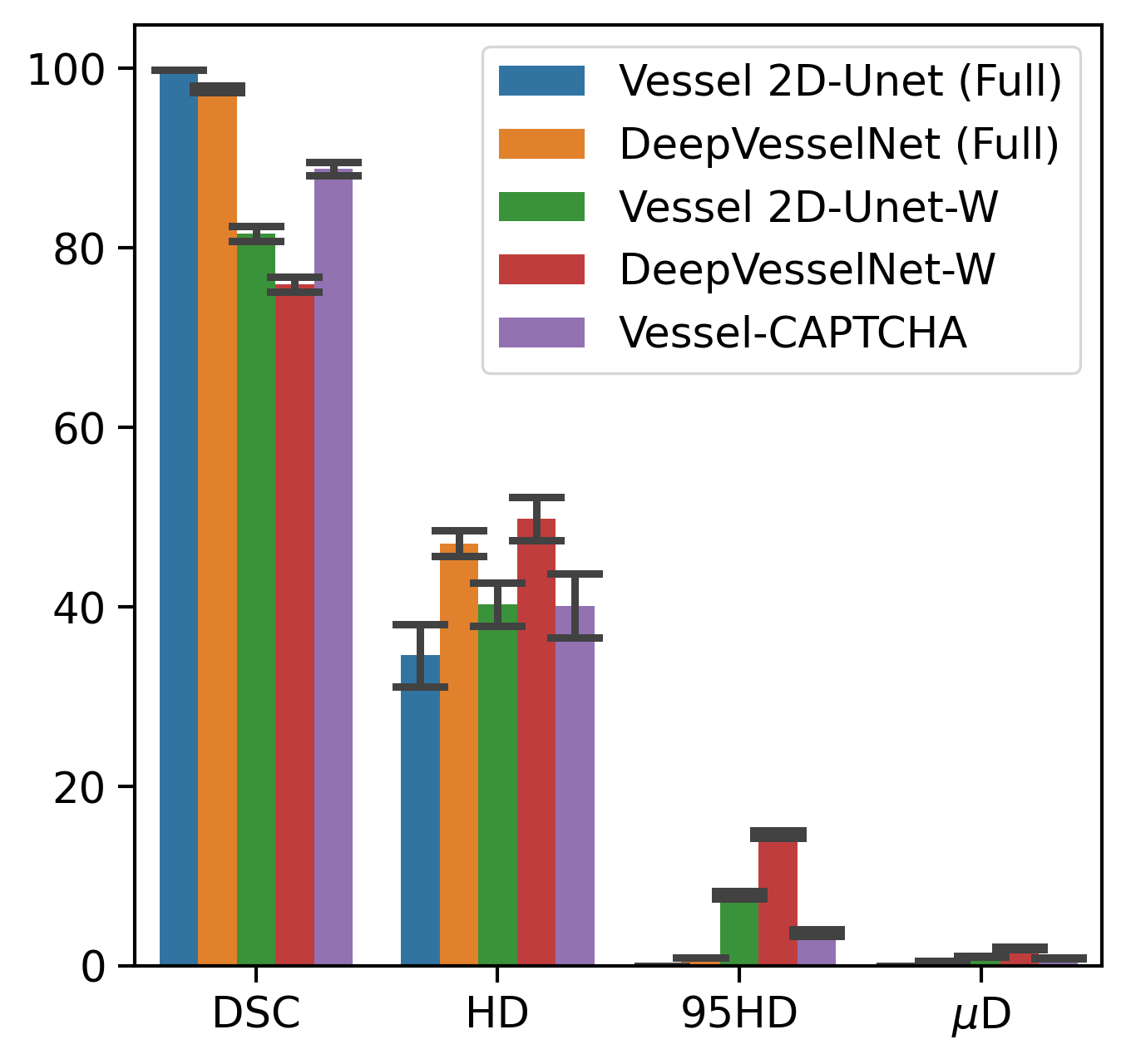} }
	\caption{\toadd{Segmentation performance in synthetic data. Vessel 2D-Unet and DeepVesselNet are trained with full pixel-wise annotations (Full) and with weak labels (-W). A higher value is better for DSC, lower is better for HD, 95HD and $\mu$D.}}
	\label{fig:synthetic}
\end{figure}

\toadd{
	\paragraph{Synthetic Data} 
	We use the synthetic data to provide a controlled setup, where the ground truth is fully reliable, to assess the learning-based vessel segmentation strategies. In addition to the required fully supervised training, Vessel 2D-Unet and DeepVesselNet are trained using weak labels from the Vessel-}\verb|CAPTCHA| \toadd{annotation scheme. We denote them as Vessel 2D-Unet-W and DeepVesselNet-W. 
	
	Figure~\ref{fig:synthetic} summarizes the segmentation accuracy results from the different networks. The Vessel 2D-Unet and DeepVesselNet present the best performances when they are trained using fully labeled and reliable ground truth data. DeepVesselNet reports a minor drop in performance ($1-2$\%) w.r.t. the values reported in \citep{tetteh2020}, which we consider related to implementation details.
	As it could be expected, the Vessel-}\verb|CAPTCHA| \toadd{has a slightly lower performance than Vessel 2D-Unet and DeepVesselNet trained with full precision labels. However, it surpasses the performance of Vessel 2D-Unet-W and DeepVesselNet-W, the same architectures trained with weak labels, indicating that  Vessel-}\verb|CAPTCHA| \toadd{is better suited for the weak learning setup.}

\toadd{
	\paragraph{TOF images}
	W}e use real clinical data from the TOF images to evaluate the Vessel-\verb|CAPTCHA| and to compare it against the 3D vessel segmentation baselines in terms of segmentation accuracy and training set annotation time. 

\toadd{Among classical 3D vessel segmentation methods,} the Frangi \toadd{\citep{Frangi1998}} and \toadd{Sato \citep{Sato1997}} filters produce real-valued maps that need to be thresholded to get a binary segmentation. The \toadd{TV \citep{Zuluaga2014a}} provides a probability map, which may produce small spurious segmentations that need to be filtered out. \toadd{The three methods allow to identify vessels at different spatial resolutions. In our experiments, we set 10 scales in the range $[0.5,2]$ mm.} We obtain final binary segmentations for the \toadd{classical} methods in two ways: 
\begin{enumerate}
	\item \toadd{\textbf{No post-processing (NP):}} the real-valued masks obtained with the Frangi and Sato filter are normalized to the range $[0,1]$. We set a fixed threshold ($t>0.6$) to binarize the three maps, and we do no filter out potential small spurious objects.
	\item \toadd{\textbf{Post-processing (PP):}} Every (real-valued and probability) map is inspected by overlaying it on the original testing image, to define and apply a per-image threshold. The resulting binary maps are filtered by masking out any connected component with a size equal or smaller than 4. Through visual inspection of every binary segmentation overlaid in the original image, the minimum connected component size could be modified. \toadd{Where the results are yet not satisfactory, the base method can be re-run using a different set of scales, followed by a new round of post-processing operations. We record the time required to obtain a visually satisfactory segmentation.}
\end{enumerate}

\begin{table*}
	\caption{3D brain vessel segmentation methods accuracy in TOF. The bold font denotes best value, with underlined values not significantly different from it ($\alpha=0.05)$. \toadd{Classical methods and DeepVesselNet use 3D volumes as input. Vessel 2D-Unet and our framework use 2D patches as inputs. HD, 95HD and $\mu$D are reported in voxels}.}
	\setlength{\tabcolsep}{5pt}
	\centering
	\begin{tabular}{|l|c|c|c|c|}
		\hline
		Method & DSC \toadd{($\uparrow$)} & \toadd{HD ($\downarrow$)} & \toadd{95HD ($\downarrow$)} & \toadd{$\mu$D ($\downarrow$)} \\
		\hline
		Frangi-NP& \multicolumn{1}{r|}{\toadd{54.16$\pm$8.81}}&\multicolumn{1}{r|}{\toadd{81.04$\pm$18.48}}&\multicolumn{1}{r|}{\toadd{14.78$\pm$13.83}}& \multicolumn{1}{r|}{\toadd{2.47$\pm$2.22}}\\
		Sato-NP& \multicolumn{1}{r|}{\toadd{55.75$\pm$7.15}}&\multicolumn{1}{r|}{\toadd{78.60$\pm$16.37}}&\multicolumn{1}{r|}{\toadd{11.53$\pm$12.01}}& \multicolumn{1}{r|}{\toadd{2.17$\pm$1.07}}\\
		TV-NP & \multicolumn{1}{r|}{\toadd{68.41$\pm$5.01}}&\multicolumn{1}{r|}{\toadd{60.23$\pm$10.08}}&\multicolumn{1}{r|}{\toadd{10.97$\pm$11.72}}& \multicolumn{1}{r|}{\toadd{2.10$\pm$1.00}}\\ 
		Frangi-PP& \multicolumn{1}{r|}{\toadd{68.44$\pm$3.15}}&\multicolumn{1}{r|}{\toadd{\underline{20.60$\pm$10.91}}}&\multicolumn{1}{r|}{\toadd{9.01$\pm$10.38}}& \multicolumn{1}{r|}{{\toadd{2.36$\pm$2.01}}}\\
		Sato-PP& \multicolumn{1}{r|}{\toadd{69.01$\pm$3.67}}&\multicolumn{1}{r|}{\toadd{\underline{21.53$\pm$9.11}}}&\multicolumn{1}{r|}{\toadd{8.86$\pm$10.09}}& \multicolumn{1}{r|}{{\toadd{2.10$\pm$1.01}}}\\
		TV-PP & \multicolumn{1}{r|}{\toadd{70.74$\pm$3.38}}&\multicolumn{1}{r|}{\toadd{\textbf{20.11$\pm$8.45}}}&\multicolumn{1}{r|}{\toadd{8.31$\pm$8.23}}& \multicolumn{1}{r|}{{\toadd{2.07$\pm$1.02}}}\\
		Vessel 2D-Unet& \multicolumn{1}{r|}{\toadd{\underline{77.66$\pm$4.32}}}&\multicolumn{1}{r|}{\toadd{74.78$\pm$16.73}}&\multicolumn{1}{r|}{\toadd{12.60$\pm$18.16}}& \multicolumn{1}{r|}{\toadd{\underline{0.60$\pm$0.11}}}\\
		\toadd{DeepVesselNet}& \multicolumn{1}{r|}{\toadd{\underline{76.13$\pm$5.51}}}&\multicolumn{1}{r|}{\toadd{75.32$\pm$12.94}}&\multicolumn{1}{r|}{\toadd{\underline{4.32$\pm$1.16}}}& \multicolumn{1}{r|}{\toadd{1.65$\pm$0.26}}\\
		Vessel-\verb|CAPTCHA|& \multicolumn{1}{r|}{\toadd{\textbf{79.32$\pm$3.02}}}&\multicolumn{1}{r|}{\toadd{51.70$\pm$5.92}}&\multicolumn{1}{r|}{\toadd{\textbf{4.06$\pm$1.50}}}& \multicolumn{1}{r|}{\toadd{\textbf{0.50$\pm$0.09}}}\\
		\hline
		\multicolumn{3}{l}{NP: No post-processing, PP: Post-processing}\\
		
	\end{tabular}
	\label{tab:all_comparison}
\end{table*}

\begin{figure}
	\centerline{
		\includegraphics[width=\columnwidth]{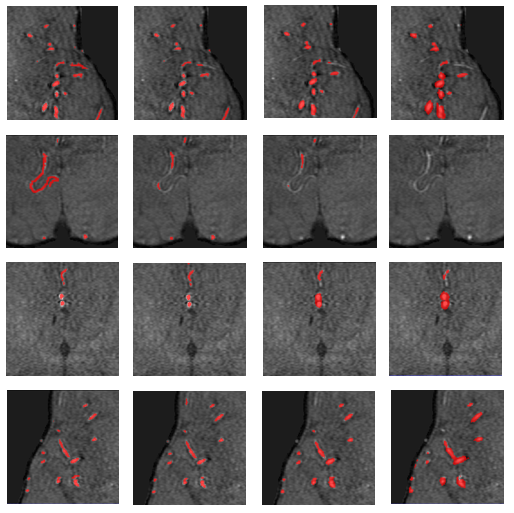} 
	}
	\caption{Segmentation results in TOF images. From left to right: ground truth, Vessel-CAPTCHA, Vessel 2D-Unet \toadd{and DeepVesselNet.}}
	\label{fig:tof}
\end{figure}

Table~\ref{tab:all_comparison} summarizes the segmentation performance. Classical vessel segmentation methods show a poor performance when no manual post-processing is done. This is expected, as it is a well-known limitation of such approaches. The manual post-processing step allows an important jump in performance. \toadd{In particular, it allows to remove spurious and disconnected false positives, which is reflected on their low HD, the best among all methods, and an important drop of the 95HD, while maintaining $\mu$D relatively constant}. However, post-processing requires high level of expertise and it is time consuming. 

\toadd{With the exception of the HD, learning-based methods consistently show a better performance across measures, with no statistical differences among them, and} the Vessel-\verb|CAPTCHA| reporting the best results among all methods. This demonstrates that the proposed framework can reach state-of-the-art performance despite the use of less accurate annotations (Fig.~\ref{fig:tof}). \toadd{We bring attention to the fact that Vessel 2D-Unet and DeepVesselNet report lower DSC (77.66 vs. 89.0 and 76.13 vs. 81.0, respectively) than the reported in \citep{Livne2019,tetteh2020}. However, for Vessel 2D-Unet our results show a better 95HD (12.6 vs 47.27) and a comparable sub-voxel $\mu$D (0.60 vs 0.38). The better distance-based measures suggest that the differences in the DSC might come from the ground truth annotation protocol, in which our data might include more distal, hence thinner vessels that are more prone to be unsegmented. This is confirmed by DeepVesselNet's DSC on synthetic data. In the controlled setup, the reported results are comparable to \citep{tetteh2020}.}


%

\begin{figure}[t]
	\centerline{
		\includegraphics[width=0.8\columnwidth]{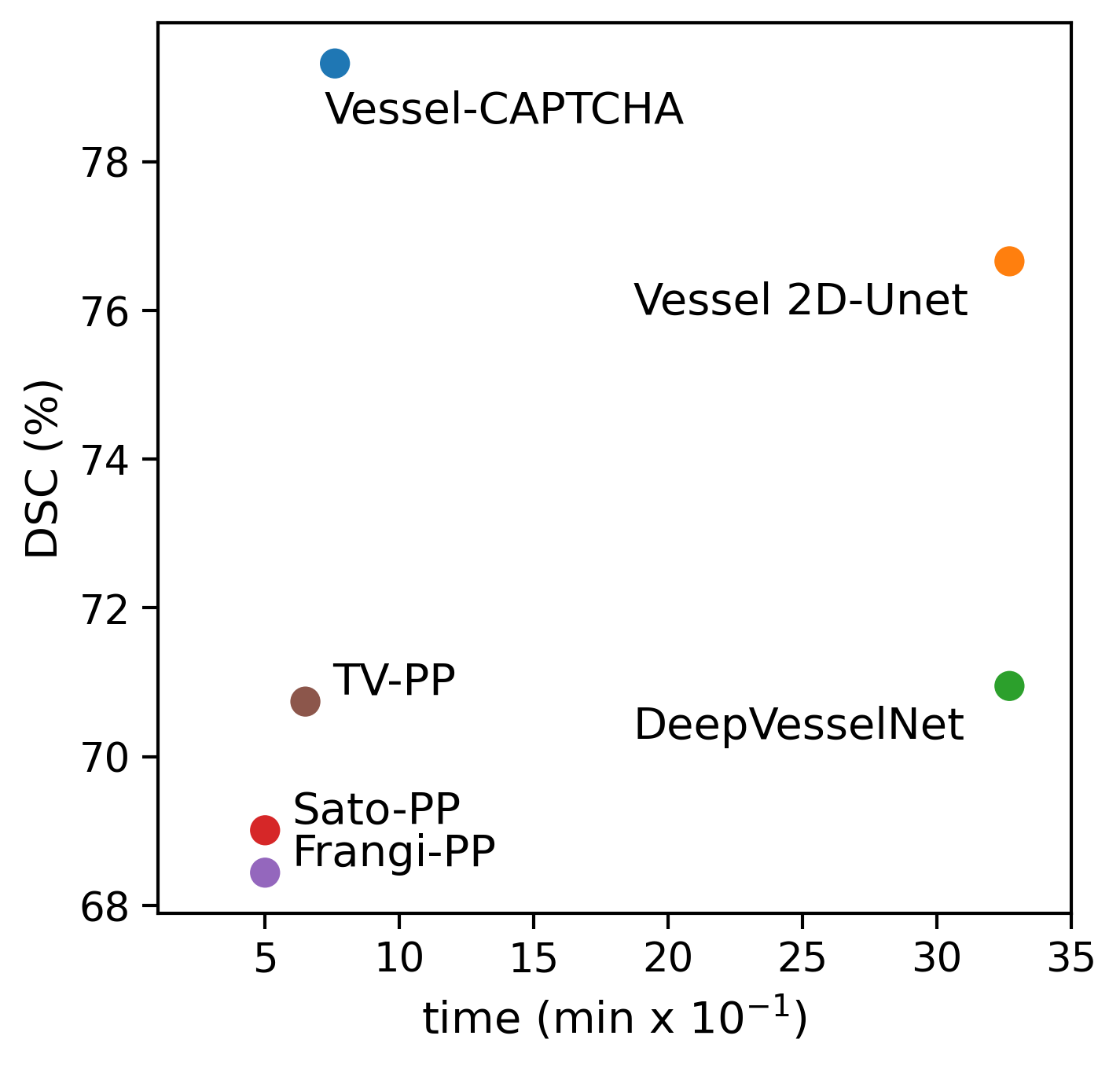} }
	\caption{\toadd{Segmentation accuracy measured with the DSC vs. User intervention time.}}
	\label{fig:times}
\end{figure}

\toadd{
	Figure~\ref{fig:times} presents segmentation accuracy measured with the DSC as a function of the required average user intervention time per image. For the proposed framework, the user intervention time corresponds to the average time required to obtain weak labels using the Vessel-}\verb|CAPTCHA| \toadd{annotation scheme. We report the average from the time measurements from the three raters (75.5$\pm$12.5 min). For 2D Vessel-Unet and DeepVesselNet, the user intervention time corresponds to the average time to fully pixel-wise annotate TOF images (327.5$\pm$20.5 min).} The 2D-Unet framework \citep{Livne2019} requires additional data pre-processing to obtain patches with vessels located at the center of the patch, which is not considered in the reported numbers. While this operation could represent a further increase in the time needed to prepare the training set, \toadd{we consider it marginal in comparison with the time required to do the pixel-wise annotation. Finally, for the classical methods, the user intervention time corresponds to the average time required to segment and post-process one image.} We observe that,  \toadd{on average}, the Vessel-\verb|CAPTCHA| reduces the annotation time by \toadd{77\%}, w.r.t. pixel-wise annotations in the same image, while achieving a higher segmentation accuracy. 

\toadd{
	\paragraph{Susceptibility-Weighted Images (SWI)} W}e study the capacity of the Vessel-\verb|CAPTCHA| \toadd{to segment different image modalities} by qualitatively assessing the segmentation results obtained in SWI. The framework was trained and visually assessed on the validation set. The model visually judged as best was used to segment the test set.

\begin{figure}
	\centerline{
		\includegraphics[width=\columnwidth]{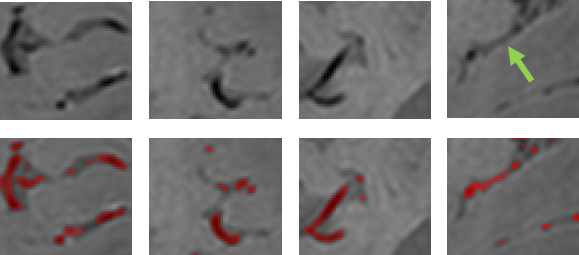} }
	\caption{Segmentation results in SWI images. Top: Original image. Bottom: Overlaid segmentation. From left to right the first three cases present good segmentation results. The rightmost example shows a sulci that has been segmented as if it was a vessel (green arrow).}
	\label{fig:swi}
\end{figure}

\toadd{Figure~\ref{fig:swi} illustrates some segmentation results.} Overall, SWI is more complex than TOF, thus further errors are observed. As a general pattern, the SWI segmentations tend to miss small vessels, while there is also a high incidence of false positives due to erroneously segmented sulci and noise. \toadd{Nevertheless, the raters judged more that 50\% of the segmentations as good and only one image was considered poor by one of them. Their visual judgment an average rating score of 2.57 with an agreement $\kappa$=0.75.}  


SWI Vessel-\verb|CAPTCHA| annotation requires 38\% more time than in TOF (94.5$\pm$11.5). This is expected given the increased complexity of SWI scans: small vessels require more effort to be identified and vessels often present an appearance similar to sulci (Fig~\ref{fig:swi}). These factors have a direct incidence in the time needed by a rater to discriminate vessel from non-vessel patches. Nevertheless, SWI Vessel-\verb|CAPTCHA| accounts for 71\% less time than the pixel-wise annotation baseline \toadd{(327.5$\pm$20.5 min, see Fig.~\ref{fig:times})}.

\toadd{
	\subsection{Alternative Limited Supervision Strategies}\label{subsec:weak}
	Using the TOF dataset, we choose to do a separate comparison of the Vessel-}\verb|CAPTCHA| \toadd{and other limited supervision strategies, which excludes fully supervised 3D brain vessel segmentation approaches. As there are no works using limited supervision addressing 3D brain vessel segmentation we consider that a direct comparison between the two families of methods is advantageous towards the fully supervised techniques.   
	
	\paragraph{Partial Labeling Techniques} Table~\ref{tab:weak_comparison} compares our framework with the partial labeling techniques, 3D-Unet, and Pseudo-labeling. The 3D-Unet is trained with the pixel-wise annotations, under the assumption that these are highly prone to error, given the difficulties that the brain vessel tree poses for annotation. Pseudo-labeling uses rough segmentation masks obtained using the Sato filter \citep{Sato1997} to the image volumes, thus avoiding user annotations}. Despite being designed to handle scarse pixel-wise annotations and being the only method directly processing the image volume, the 3D-Unet does not achieve the best performance. \toadd{The results are lower than those reported by other frameworks requiring precise pixel-wise annotations, i.e. Vessel 2D-Unet and DeepVesselNet (Table~\ref{tab:all_comparison}). These results are consistent with other works in the literature  \citep{Livne2019,kozinski2020,Ni2020,Phellan2017,tetteh2020}, which avoid the use of end-to-end 3D networks and favor the use of networks relying on smaller input spaces, e.g. 3D subvolumes \citep{Phellan2017,tetteh2020}, 2D images \citep{kozinski2020,Ni2020} or patches~\citep{Livne2019}. Pseudo-labeling results suggest that, in isolation, this approach cannot reach a good accuracy, which explains why it is often coupled with a refinement stage \citep{Liang2019,Ke2020}.  
}

\begin{table}
	\caption{\toadd{Comparison with partial labeling methods using TOF. The bold font denotes best value. Our framework uses 2D patches, Pseudo-labeling image slices and 3D-Unet image volumes as input. 
	}}
	\setlength{\tabcolsep}{5pt}
	\centering
	\small
	\begin{tabular}{|l|c|c|c|}
		\hline
		& 3D-Unet & \toadd{Pseudo-labeling} & Vessel-\verb|CAPTCHA|\\
		\hline
		DSC \toadd{($\uparrow$)} & \multicolumn{1}{r|}{\toadd{68.50$\pm$3.37}} & \multicolumn{1}{r|}{\toadd{54.99$\pm$5.86}}& \multicolumn{1}{r|}{\toadd{\textbf{79.32$\pm$3.02}}}\\ 
		\toadd{HD ($\downarrow$)} & \multicolumn{1}{r|}{\toadd{76.12$\pm$8.47}} &\multicolumn{1}{r|}{\toadd{68.50$\pm$9.58}} &\multicolumn{1}{r|}{\toadd{\textbf{51.70$\pm$5.92}}} \\
		\toadd{95HD ($\downarrow$)} & \multicolumn{1}{r|}{\toadd{15.72$\pm$2.23}} &\multicolumn{1}{r|}{\toadd{24.19$\pm$5.25}}&\multicolumn{1}{r|}{\toadd{\textbf{4.06$\pm$1.50}}}\\
		\toadd{$\mu$D ($\downarrow$)} & \multicolumn{1}{r|}{\toadd{2.56$\pm$1.44}}& \multicolumn{1}{r|}{\toadd{4.48$\pm$1.67}} &\multicolumn{1}{r|}{\toadd{\textbf{0.50$\pm$0.09}}}\\
		\hline
	\end{tabular}
	\label{tab:weak_comparison}
\end{table}

\toadd{
	\paragraph{Weakly Supervised Strategies} In our experiments, we were not able to achieve sufficiently good results with WS-MIL and AffinityNet that could allow a quantitative comparison with the other baselines. In this section, we perform a qualitative analysis of the obtained results to gain understanding about the limitations of standard MIL- and CAM-based segmentation techniques for brain vessel tree segmentation.  
	
	We adapt WS-MIL to address 3D brain vessel segmentation by using the Vessel-}\verb|CAPTCHA| \toadd{patches as input rather than an image slice \citep{Lerousseau2020}. WS-MIL splits its input into sub-patches and it ranks them according to their predicted probability of containing a vessel. We consider two sub-patch sizes, 16$\times$16 and 8$\times$8. The final sub-patch labeling is achieved by using the ranked patches along with two hyper-parameters, $\alpha$ and $\beta$, which control the minimum number of pixels belonging to the foreground ($\alpha$) and the background class ($\beta$) (Table~\ref{tab:others}). We observe two limitations in the obtained results (Fig.~\ref{fig:mils}). First, the resulting masks correspond to vessel localization masks, not segmentations, due to the granularity of the patches. The original WS-MIL formulation \citep{Lerousseau2020} has been conceived for super resolution histology images, where the resulting labeled sub-patches can be considered a segmentation mask. Standard brain images have a much lower resolution. Therefore, the final result lacks the necessary specificity to be considered a segmentation. Second, we observe that it is difficult to set a value for $\alpha$ and $\beta$ that works well for all the slices in an image volume. As shown in Figure~\ref{fig:mils}, while a low $\alpha$ value works well in image slices with larger vessels, the same value fails to detect smaller vessels, hence it is necessary to train a new model with different $\alpha$, $\beta$ values. 
	
	\begin{figure}[t]
		\centerline{
			\includegraphics[width=\columnwidth]{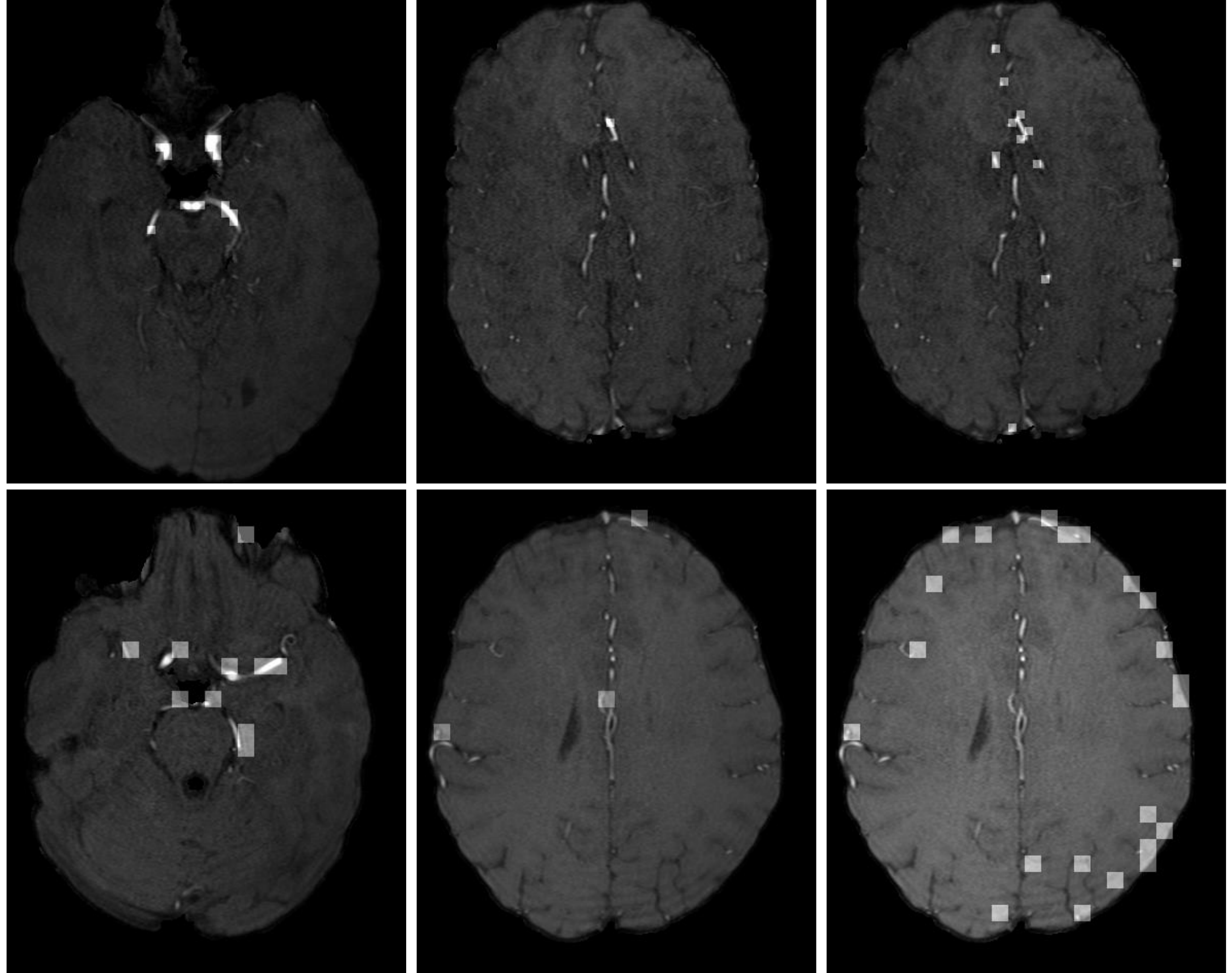} }
		\caption{\toadd{Vessel localization results with WS-MIL using sub-patch resolution 8$\times$8 (top) and 16$\times$16 (bottom). The first two columns use $\alpha=0.01$, $\beta=0.99$. The right-most column uses $\alpha=0.07$, $\beta=0.93$ on the middle column images.}}
		\label{fig:mils}
	\end{figure}
	
	\begin{figure}
		\centerline{
			\includegraphics[width=\columnwidth]{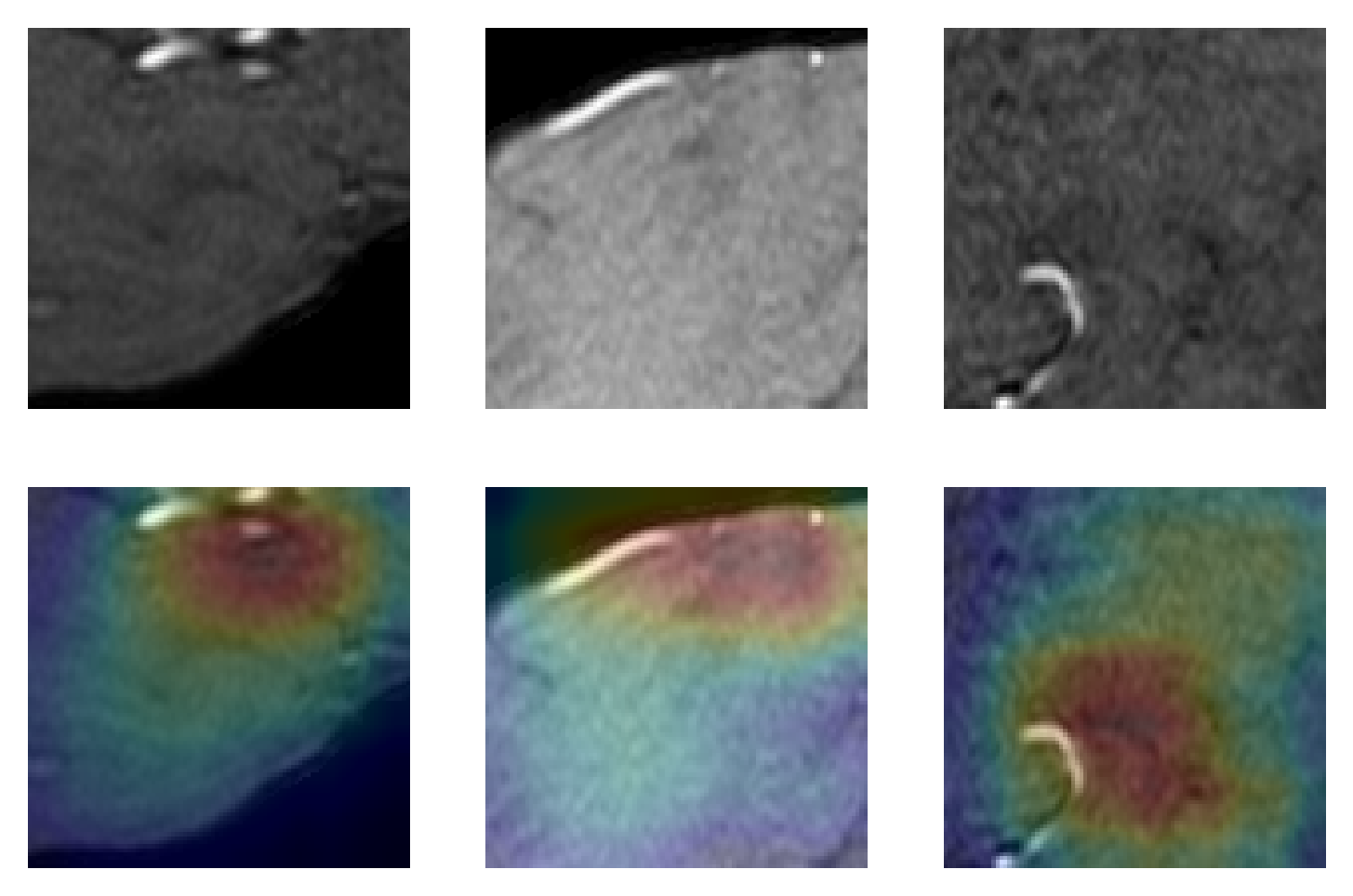} }
		\caption{\toadd{Vessel patches of size 96$\times$96 (top) with overlaid CAMs (bottom) from the AffinityNet framework.}}
		\label{fig:cams}
	\end{figure}
	
	The architecture of AffinityNet does not allow images below a certain size to be fed into it. Therefore, we had to enlarge the patch used from 32$\times$32 to 96$\times$96, similar to the one we use as input of 2D-WnetSeg. The larger patches were obtained by grouping 32$\times$32 patches. A vessel label was assigned if at least one sub-patch was originally labeled as a vessel patch. Otherwise, the patch was labeled as non-vessel. 
	
	Despite the larger field of view of the new input patches, our experiments did not achieve good results with AffinityNet. A visual inspection of the CAMs showed that, although they activate consequently with the class associated to the patch, these did not contain discriminative information about vessels (Fig.~\ref{fig:cams}).  
	Let us recall that AffinityNet \citep{Ahn2018} uses the input image and the CAMs \citep{Zhou2016} to synthesize pseudo-labels, which are then used to train a segmentation model. However, CAMs are rough approximations of the object of interest \citep{Ahn2018,Bae2020,Zou2021}. In the past, CAM-based methods have been used to segment relatively large objects in natural scenes \citep{Ahn2018,Hong2017,Zou2021}, damaged tissue \citep{izadyyazdanabadi2018} or blob-like structures occupying an important part of the image, such as the optic disc \citep{Zhao2019}. In our case, as vessels are relatively small objects, it seems that the network requires to use much more information from the scene to discriminate between vessel and non-vessel patches, as reflected by the CAMs (Fig.~\ref{fig:cams}). The information, however, is to broad to locate the vessels and thus AffinityNet fails.

}

\begin{table*}[t]
	\centering
	\caption{Classification network comparison in TOF and SWI. For each row, bold font denotes the best value, with underlined values not significantly different from it ($\alpha=0.05)$}
	
	\begin{tabular}{|p{15pt}|p{37pt}|p{60pt}|p{60pt}|p{60pt}|p{60pt}|}
		\hline
		
		\multicolumn{2}{|r|}{} & \multicolumn{1}{c|}{VGG-16} & \multicolumn{1}{c|}{ResNet} & \multicolumn{1}{c|}{2D-UnetCl} & \multicolumn{1}{c|}{2D-PnetCl}\\
		\hline 
		& \multicolumn{1}{c|}{Precision} 
		& \multicolumn{1}{r|}{92.48$\pm$1.54} 
		& \multicolumn{1}{r|}{93.66$\pm$1.48} 
		& \multicolumn{1}{r|}{\underline{94.82$\pm$0.48}} 
		& \multicolumn{1}{r|}{\textbf{94.91$\pm$1.04}} \\
		\multicolumn{1}{|c|}{TOF} & \multicolumn{1}{c|}{Recall} 
		& \multicolumn{1}{r|}{87.39$\pm$4.60} 
		& \multicolumn{1}{r|}{93.27$\pm$1.73} 
		& \multicolumn{1}{r|}{\underline{94.04$\pm$0.65}} 
		& \multicolumn{1}{r|}{\textbf{94.94$\pm$1.09}}\\	
		& \multicolumn{1}{c|}{F-score} 
		& \multicolumn{1}{r|}{88.68$\pm$3.81} 
		& \multicolumn{1}{r|}{93.34$\pm$1.62} 
		& \multicolumn{1}{r|}{\underline{94.27$\pm$0.54}}
		& \multicolumn{1}{r|}{\textbf{94.71$\pm$1.23}}\\
		\hline
		& \multicolumn{1}{c|}{Precision} 
		& \multicolumn{1}{r|}{\underline{82.34$\pm$1.15}} 
		& \multicolumn{1}{r|}{80.14$\pm$1.13} 
		& \multicolumn{1}{r|}{\underline{82.44$\pm$1.18}} 
		& \multicolumn{1}{r|}{\textbf{82.97$\pm$1.55}} \\
		\multicolumn{1}{|c|}{SWI} 
		& \multicolumn{1}{c|}{Recall} 
		& \multicolumn{1}{r|}{77.45$\pm$4.17} 
		& \multicolumn{1}{r|}{\textbf{79.39$\pm$3.35}} 
		& \multicolumn{1}{r|}{74.35$\pm$5.35} 
		& \multicolumn{1}{r|}{\underline{79.30$\pm$4.07}}\\	
		& \multicolumn{1}{c|}{F-score} 
		& \multicolumn{1}{r|}{78.76$\pm$3.39} 
		& \multicolumn{1}{r|}{\underline{79.17$\pm$2.31}} 
		& \multicolumn{1}{r|}{76.42$\pm$4.63}
		& \multicolumn{1}{r|}{\textbf{80.31$\pm$3.31}}\\
		\hline 
	\end{tabular}
	\label{tab:classification}
\end{table*}
\toadd{
	\subsection{Classification Networks}\label{subsec:classification}}
\toadd{\paragraph{Classification Networks Performance}
	W}e study the performance of the two classification networks, 2D-UnetCl and 2D-PnetCl, to determine if they are well-suited as discriminators within our framework. 
Table~\ref{tab:classification} compares the classification performance of 2D-UnetCl and 2D-PnetCl in TOF and SWI images with VGG-16 and the ResNet. For each network, two models were trained, one for TOF and one for SWI.
Results are reported on the best performing model in the validation set. 

The two proposed networks, derived from medical imaging task-specific networks, present a higher overall performance (F-score) than VGG-16 and the ResNet, suggesting that the networks specifically designed for medical imaging tasks can contribute to an increased performance. 
All methods report a drop in performance from TOF to SWI, which is expected given that SWIs are more challenging to classify and segment due to several factors. First, vessels in SWI are hypo-intense, being similar in appearance to the image background. As such, vessels close to the brain surface are prone to misclassification. Second, SWI is capable of imaging very small vessels that can be difficult to identify within a patch, as they can have an appearance similar to the one of brain tissue inhomogeneities or sulci, this leading to misclassification. 

Among the proposed networks, 2D-PnetCl presents the highest performance in both modalities. This reflects a good balance in the network's capability to discriminate among vessel and non-vessel patches, which is key for its use within the Vessel-\verb|CAPTCHA| framework. In the remaining, we rely on 2D-PnetCl as a classification network. 

\begin{figure}[t]
	\centerline{
		\includegraphics[width=0.8
		\columnwidth]{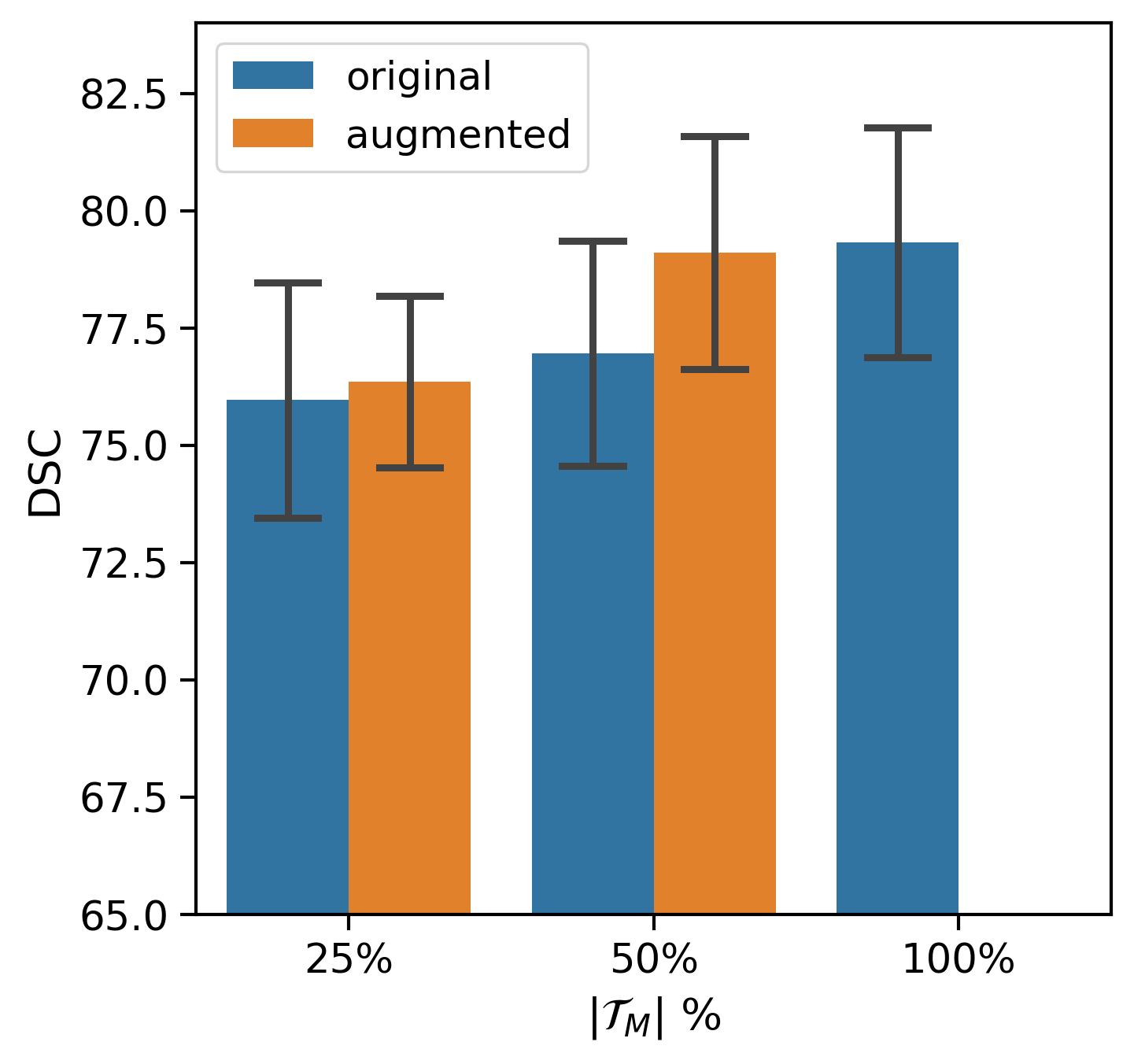} 
	}
	\caption{\toadd{Segmentation performance with varying training set size with (augmented) and without (original) data augmentation.}}
	\label{fig:augmented}
\end{figure}

\toadd{
	\paragraph{Classification Network as a Weak Pseudo-label Generator}
	We use a percentage (25\%, 50\% and 100\%) of the weakly annotated training set $\mathcal{T}_M$. Where applicable, we enlarge it with a fixed set of 10 images automatically labeled through the data augmentation process, i.e. $|\mathcal{T}_M^*|$=10, (Fig.~\ref{fig:trainingplus}).}
Figure~\ref{fig:augmented} reports DSC in the different scenarios.
The results show that the data augmentation step improves performance w.r.t. using the same annotated training set with no augmentation, while reaching a comparable performance to that one of using a dataset entirely annotated by the user. The comparable performances come as a result of the high classification accuracy of the 2D-PnetCl (F-score=94.71\%), which sits close to the performance of a human rater. 


\begin{figure}[t]
	\centerline{
		\includegraphics[width=0.95\columnwidth]{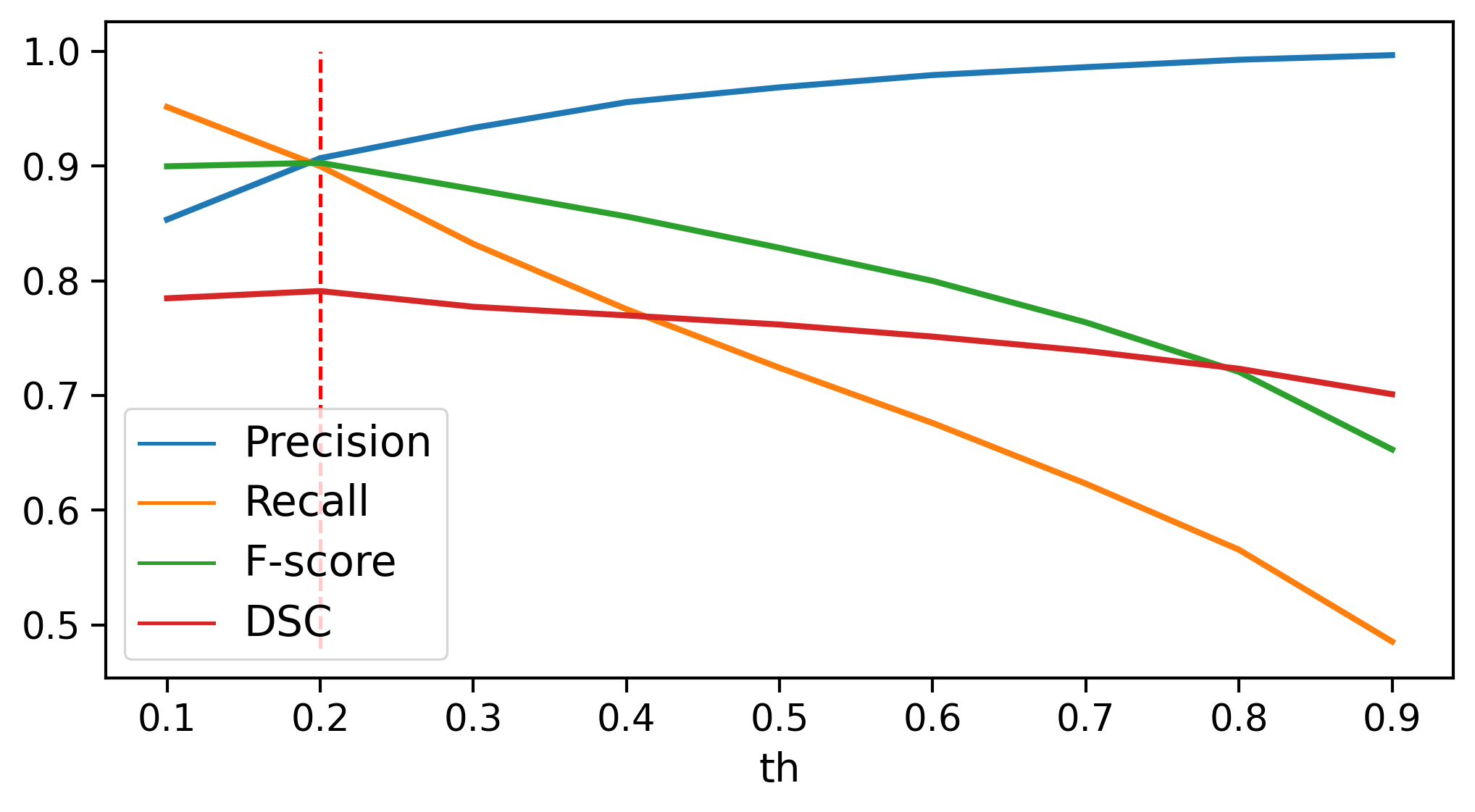} 
	}
	\caption{\toadd{Threshold (th) calibration of the 2D-PnetCl output.}}
	\label{fig:class_thresh}
\end{figure}

\toadd{
	\paragraph{Classification Network as a Second Opinion}
	The results obtained by post-processed classical methods (Table~\ref{tab:all_comparison}) suggest that a revision of the  segmentation results and their refinement through post-processing can lead to a significant improvement in performance. We investigate if the classification network can act as an expert providing a second opinion on the segmentation results obtained by the 2D-WnetSeg, on a per-patch basis. If the classification network labels a patch as vessel patch, the segmented pixels in the patch will be preserved. Instead, if the classification network classifies the patch as a non-vessel one, any segmented pixels are masked out. To this end, we calibrate the 2D-PnetCl output by choosing the classification threshold of the final prediction layer, which maximizes the DSC (Fig.~\ref{fig:class_thresh}).} 

\begin{figure}[t]
	\centerline{
		\includegraphics[width=0.9\columnwidth]{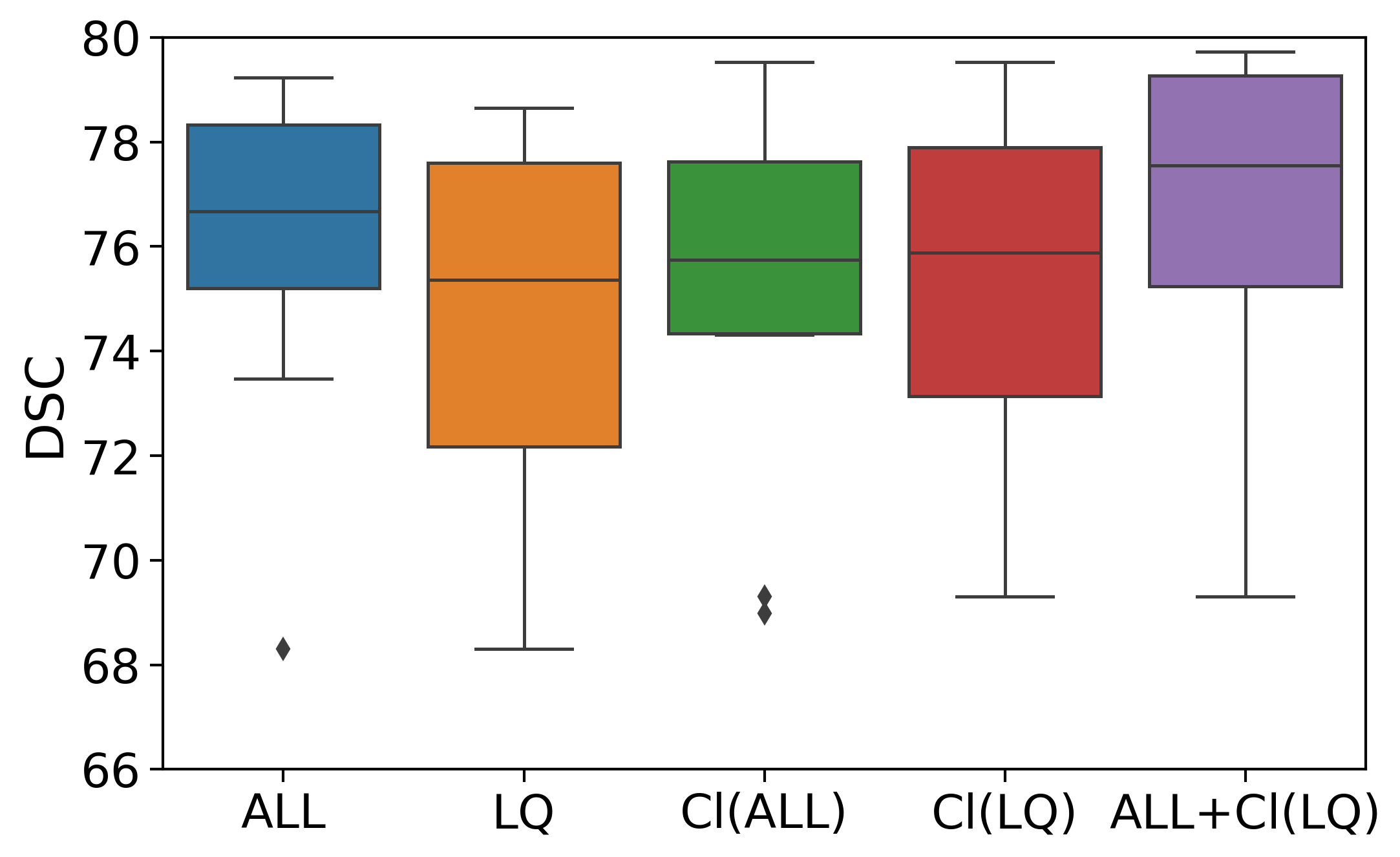} 
	}
	\caption{Classification network as a \toadd{second opinion} in TOF. Vessel segmentation DSC for all the test set (ALL), low quality test images (LQ), full test set \toadd{after second opinion} (Cl(ALL)), low quality images \toadd{after second opinion} (Cl(LQ)) and full test with only the low quality \toadd{subject to a second opinion} (ALL+Cl(LQ)) using 2D-WnetSeg trained on original training set 1.}
	\label{fig:filter}
\end{figure}

Figure~\ref{fig:filter} reports vessel segmentation DSC, \toadd{using Set 1 of the TOF images}, in the following scenarios: 1) on all the testing set (ALL); 2) on 4 images identified as of low quality (LQ); 3) \toadd{using a second opinion on} the testing set (Cl(ALL)); 4) \toadd{using a second opinion on} the low quality data (Cl(LQ)); and 5) in all the testing set with the \toadd{a second opinion} only \toadd{on the} low quality data (ALL+Cl(LQ)). 
The results suggest that \toadd{using the classifier network as a second opinion} has a significant impact in the segmentations' accuracy and variability for low quality (LQ) images ($p$-value$<$0.05), although when applied to the full test set there is a slight drop in accuracy ($\sim$1.9$\%$), indicating a negative impact on the segmentation accuracy in high quality images. As a result, one could consider \toadd{the classifier as a second opinion and not the main expert. In images were there is a discrepancy between the segmentation network and the classifier, the user may inspect them and decide what to do. As an example, the second opinion could be used only on those images } identified as of low quality. \toadd{The results from Fig~\ref{fig:filter} indicate that, in such scenario, a higher overall performance is achieved.}

\begin{figure}
	\centerline{
		\includegraphics[width=\columnwidth]{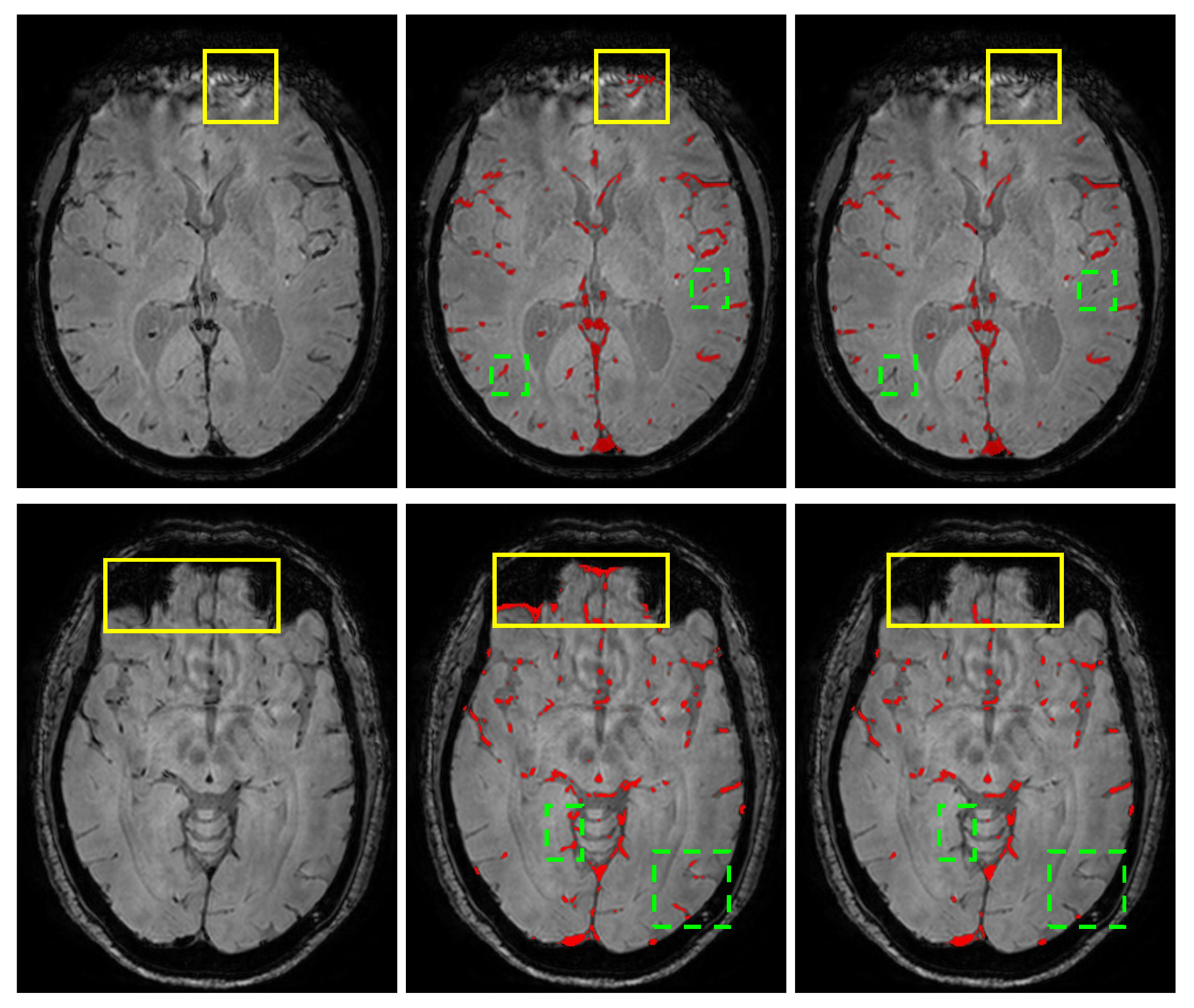} }
	\caption{Classification network as a \toadd{second expert opinion} in two SWI slices. From left to right, original image, segmentation from 2D-WnetSeg, segmentation after filtering. The yellow boxes highlight areas with image noise that are first segmented as vessel, but corrected with the filter. \toadd{The green dashed boxes, highlight areas with segmented vessels that are removed.}}
	\label{fig:swi_noise}
\end{figure}

\toadd{We follow the same procedure using SWI segmentations and present the revised segmentation masks to the raters for visual judgement. The average rating score achieved was 2.30 with an agreement $\kappa$=0.57. The lower rating score is explained by the fact that using the classification network as an expert opinion allowed to} correct segmentations containing large regions of false positives caused by noise in the image, mostly in the boundaries of the brain tissue, \toadd{at the cost of removing some true positives (Fig.~\ref{fig:swi_noise}). One rater considered this as less critical than the other, which explains the lower agreement among them. The results suggest that the classifier network should not be considered as an expert, i.e. it acts as a mask, but as a second opinion providing a heuristic measure of uncertainty on patches where the two networks disagree. The mismatching and uncertain regions should be thus validated by an external user.}




%

\subsection{Ablation Study}\label{subsec:ablation}
We study the properties of the different components of the proposed annotation and segmentation framework through a set of ablation studies. \toadd{We investigate the incidence of the K-means as and we investigate the role of the 2D-WnetSeg network.}

\toadd{
	\subsubsection{K-means as a Pseudo-label Generation Strategy}
	W}e study how the pixel-wise \toadd{pseudo-}labeled dataset $\mathcal{T}_M$ synthesized from user-provided \toadd{weak} patch tags affects the framework's performance in TOF. We achieve this in two ways. First, we investigate if the pixel-wise \toadd{pseudo-}labels synthesized by K-means represent a good rough approximation of pixel-wise user-annotated labels. Second, we assess how the size of the patches used as input of the segmentation network influences the latter's performance. In our experiments, we compare with Gaussian mixture models (GMM), an alternative self-supervised approach to obtain pixel-wise \toadd{pseudo-}labels \toadd{from image tags \citep{Luo2020}}. Two components (vessel and background) are used for the GMM to be comparable with K-means. For both cases, patches with more than 30\% pixels marked as vessel are fully masked out and considered as non-vessel. These correspond to highly noisy patches containing only brain tissue.

The role of the self-supervised method, i.e. the K-means in our case, is to synthesize pixel-wise \toadd{pseudo-}label masks $\{\mathcal{M}_s\}_{s=1}^S$ which are sufficiently good to train the segmentation network. In other words, the \toadd{pseudo-}labels should be as close as possible to hypothetically pixel-wise annotations provided by a user. We thus measure the similarity between the pixel-wise \toadd{pseudo-}labeled masks $\{\mathcal{M}_s\}_{s=1}^S$ and the available pixel-wise annotations of the TOF training set. The K-means (and GMM) are applied on different input sizes, namely directly on the full image volume, or on subsets of it that are then concatenated. For this we use image slices and patches of varying sizes: 96, 64 and 32. For the patches, K-means and GMM are only applied to vessel patches. We set 32 as the smallest patch size, which corresponds to the size set for the Vessel-\verb|CAPTCHA|, i.e. the user-input. Larger patches are obtained by concatenating the user input into a $2\times2$ and $3\times3$ grid. 

\begin{figure}[t]
	\centerline{
		\includegraphics[width=0.9\columnwidth]{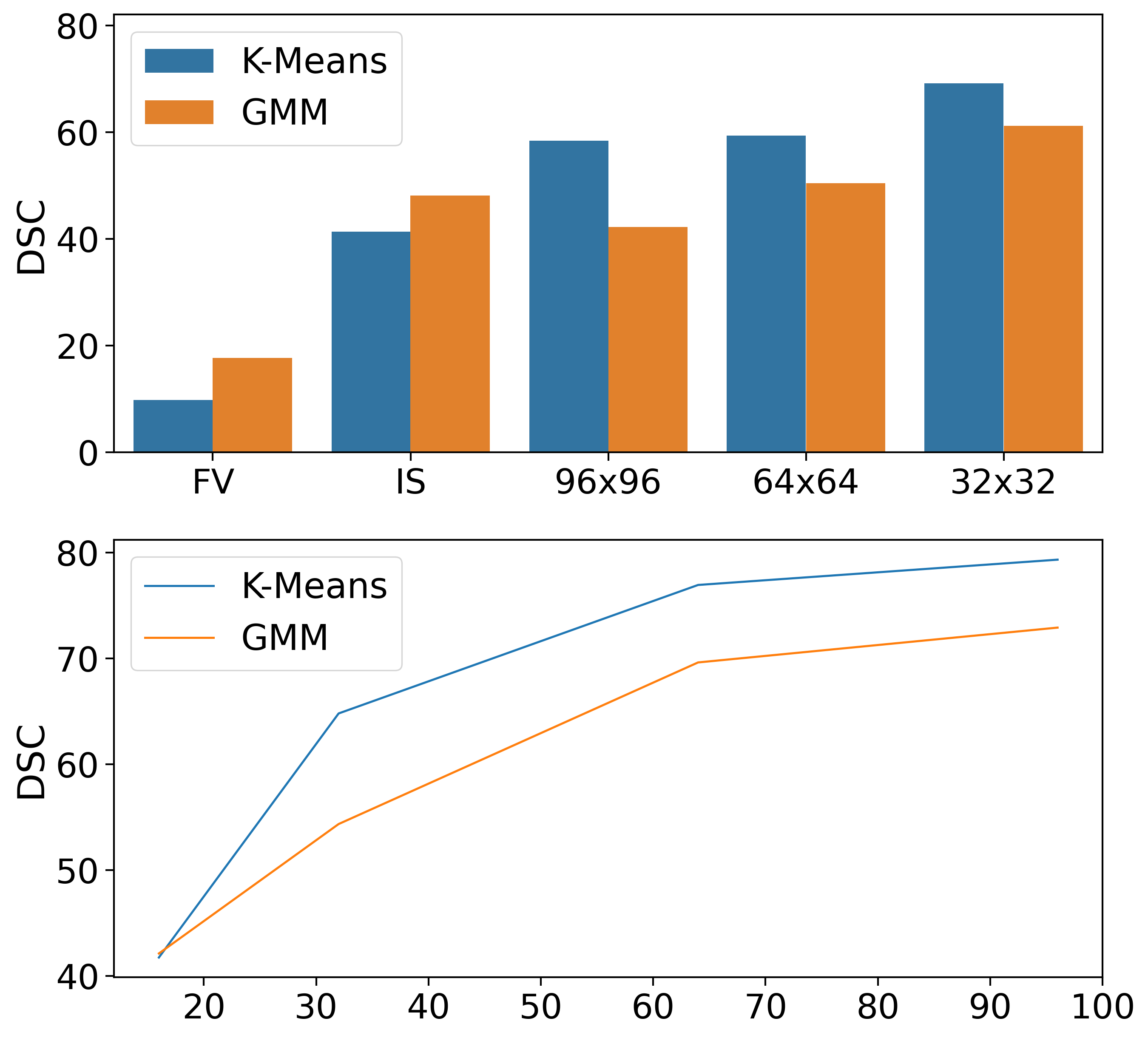} 
	}
	\caption{\toadd{Top:} Similarity between user-provided pixel-wise annotations and weak pixel-wise labels obtained through K-means and GMM, measured through the DSC in TOF. K-means and GMM are applied on the full volume (FV), on a per slice basis (IS) and on different patch sizes. \toadd{Bottom:} 2D-WnetSeg performance using pixel-wise \toadd{pseudo-}labels by K-means and GMM for different input patch sizes (16, 32, 64 and 96).}
	\label{fig:kmeansgmm_evol}
\end{figure}

\begin{figure}
	\centerline{
		\includegraphics[width=\columnwidth]{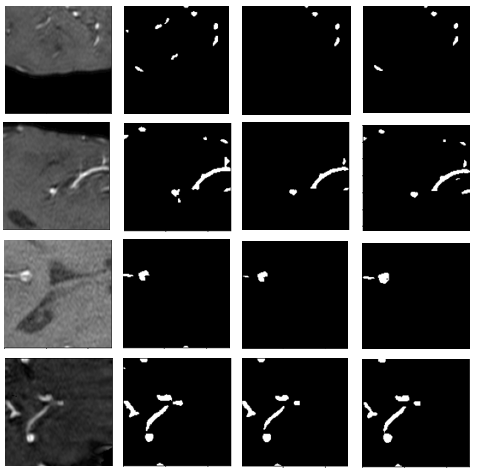} 
	}
	\caption{Examples of the generated training set $\mathcal{T}_M$. From left to right original TOF image, ground truth, GMM \toadd{pseudo-}labels and K-means \toadd{pseudo-}labels.}
	\label{fig:kmeansgmm}
\end{figure}

\toadd{\paragraph{Smaller Patches are Best for Pseudo-label Generation} F}igure~\ref{fig:kmeansgmm_evol}(top) shows the similarity between the training set pixel-wise annotations and the weak pixel-wise label masks measured with the DSC.
The performance of both methods is inverse to the size of the input sample. As it would be expected, when applied to large extents of the image volume, i.e. the full image volume (FV) or on a per image slice basis (IS), the DSC is very low ($<40\%$), with GMM reporting slightly higher values. As the extent of the input sample decreases, i.e using patches, K-means performs better, which could be justified by the fact that smaller regions tend to be more homogeneous. Two aspects should be highlighted from the obtained results. Firstly, we observe that GMMs lead to thinner vessel masks than those synthesized by K-means (Fig.~\ref{fig:kmeansgmm}), which is consistent with the higher DSC, as over-segmentations tend to be less penalized than mis-segmentations. Given the way that the 2D-WnetSeg learns, it is better to have overestimated masks from K-means than the finer ones. However, being K-means a simpler algorithm, the patch size used as the input plays an important role. Our results suggest that smaller patch sizes lead to better results. Secondly, we shall recall that both self-supervised methods are only applied to vessel patches. This is a necessary condition to obtain \toadd{pseudo-}labels of a minimum quality using these two algorithms. The condition is guaranteed by the patch tags discriminating vessel from non-vessel patches, which are obtained through the Vessel-\verb|CAPTCHA|. Based on these results, for the remaining experiments we set the patch size input to the K-means to $32\times32$, which corresponds to the same value used in the Vessel-\verb|CAPTCHA|. 

\toadd{\paragraph{Larger Patches are Best for Segmentation} F}igure~\ref{fig:kmeansgmm_evol} (bottom) shows the 2D-WnetSeg accuracy with varying input patch sizes over the validation set. The patches are obtained by rebuilding the rough mask volume from the 32$\times$32 patches and re-cropping the volume into different patch sizes. It should be noted that the segmentation network's input patch size does not have to match that one of the Vessel-\verb|CAPTCHA|. Coherently with the previous results showing that K-means \toadd{pseudo-}labels are more similar to true annotations, their use consistently leads to higher DSCs. The Vessel-\verb|CAPTCHA| patch size, $32\times32$, seems too small for the 2D-WnetSeg to capture the features that allow to discriminate vessel pixels from non-vessel ones. Instead, larger patches lead to higher DSCs. However, we avoid the use of larger patch sizes to avoid the problem of vessels becoming a small portion of the full image/patch, leading to drops in performance. For instance, we set the segmentation network's input patch size to $96\times96$. 

\begin{figure}[t]
	\centerline{
		\includegraphics[width=0.8
		\columnwidth]{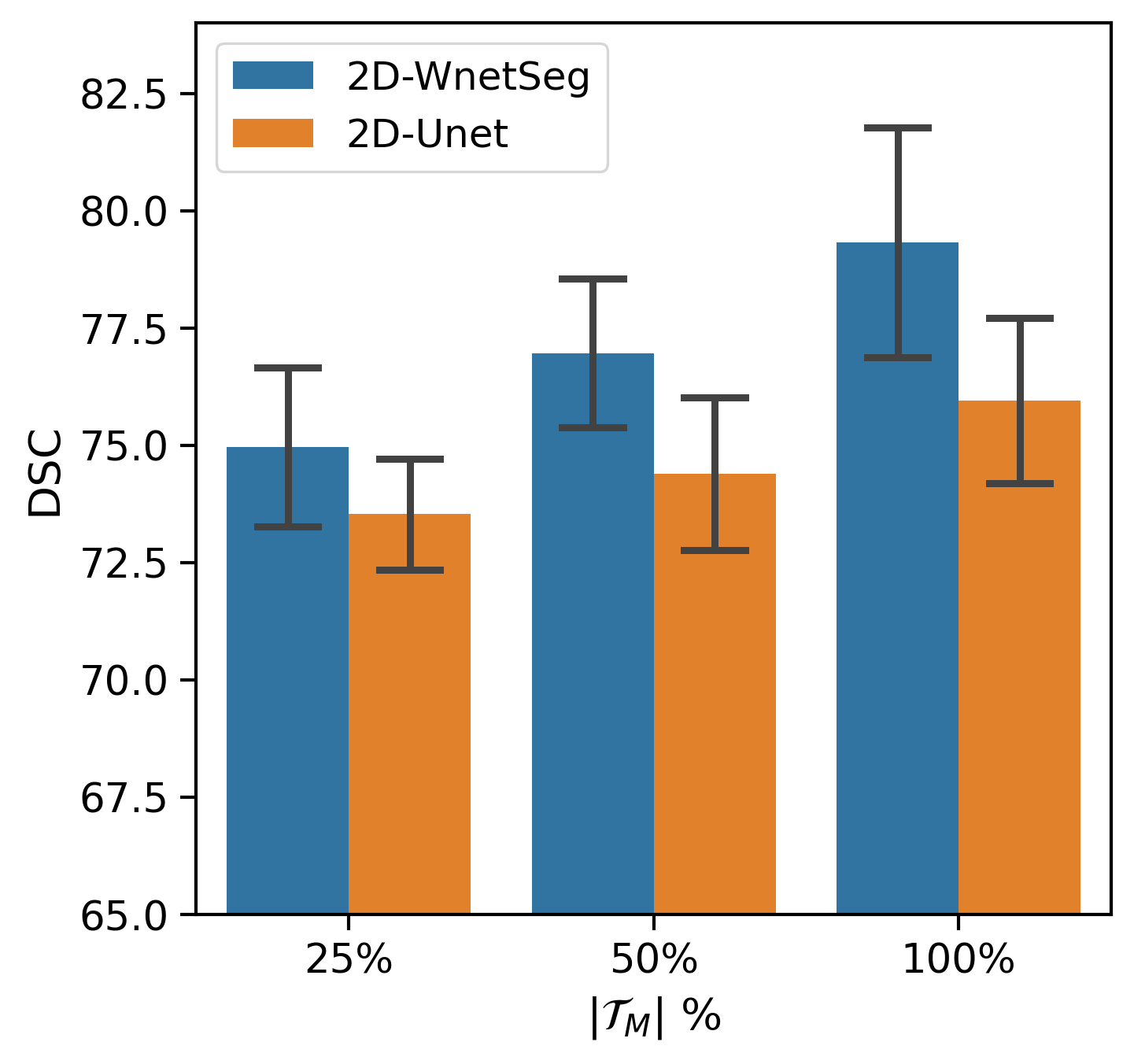} 
	}
	\caption{\toadd{2D-WnetSeg vs single Unet performance (DSC) for varying training set size,  $|\mathcal{T}_M|$}.}
	\label{fig:ablation}
\end{figure}

\toadd{
	\subsubsection{The Role of the Segmentation Network}~\label{sec:segnet}
	We perform an ablation study to explore the effectiveness of the 2D-WnetSeg}. 
Figure~\ref{fig:ablation} compares the performance of 2D-WnetSeg with its ablated version consisting its first Unet (2D-Unet), \toadd{while varying the size of the training set}. The 2D-WnetSeg reports a higher DSC across datasets. 
The better performance of the 2D-WnetSeg is explained by the fact that the deep networks are trained on rough segmentation maps. The first Unet works as a refinement module to correct the mask by inferring potentially missing vessels based on the structural redundancy of the cerebrovascular tree. The second Unet can learn from the raw brain image and the previously improved segmentation mask, leading to an increased segmentation performance. The single Unet, instead, is faced directly with the rough masks. 
\toadd{We further investigate this behavior using the synthetic dataset, which provides a controlled setup for comparison (Table~\ref{tab:ablation}). The higher reported DSC of 2D-WnetSeg indicates it is better at detecting vessel pixels. Moreover, the lower 95HD and $\mu$D are a sign of the more refined results that the 2D-WnetSeg can achieve w.r.t. its ablated version.}

\begin{table}[t]
	\caption{\toadd{2D-WnetSeg vs single Unet performance using synthetic data.}} 
	\setlength{\tabcolsep}{5pt}
	\centering
	\begin{tabular}{|p{50pt}|p{90pt}|p{90pt}|}
		\hline
		\multicolumn{1}{|c}{\toadd{Measure}}   & \multicolumn{1}{|c}{\toadd{2D-WnetSeg}} & \multicolumn{1}{|c|}{\toadd{One 2D-Unet}}\\
		\hline
		\multicolumn{1}{|l|}{\toadd{DSC ($\uparrow$)}} & \multicolumn{1}{r|}{\toadd{88.77$\pm$0.90}} & \multicolumn{1}{r|}{\toadd{86.61$\pm$1.05}} \\
		\multicolumn{1}{|l|}{\toadd{HD ($\downarrow$)}} & \multicolumn{1}{r|}{\toadd{40.31$\pm$2.95}} & \multicolumn{1}{r|}{\toadd{41.18$\pm$4.32}}\\
		\multicolumn{1}{|l|}{\toadd{95HD ($\downarrow$)}} & \multicolumn{1}{r|}{\toadd{6.74$\pm$0.48}}&\multicolumn{1}{r|}{\toadd{7.96$\pm$0.52}} \\
		\multicolumn{1}{|l|}{ \toadd{$\mu$D ($\downarrow$)}} &\multicolumn{1}{r|}{\toadd{0.91$\pm$0.06}} & \multicolumn{1}{r|}{\toadd{1.08$\pm$0.07}}\\

		\hline
	\end{tabular}
	\label{tab:ablation}
\end{table}



\begin{table*}
	\caption{Performance summary considering segmentation accuracy, model complexity (Params, GFLOPs), and computational (training and prediction) and user intervention time in minutes. In classical models (NL), user intervention time is measured during inference. In learning-based models, it refers to the time used during training set annotation. For accuracy measures, the bold font denotes best value, with underlined values not significantly different from it ($\alpha=0.05)$. }
	\small
	\setlength{\tabcolsep}{5pt}
	\centering
	\begin{tabular}{|l|l|c|c|c|c|c|c|c|c|c|}
		\hline
		&  \multirow{2}{*}{Method} & \multicolumn{4}{c|}{Accuracy} & \multicolumn{2}{c|}{\toadd{Complexity ($\downarrow$)}} & \multicolumn{3}{c|}{\toadd{Time ($\downarrow$)}}\\
		\cline{3-11}
		& & DSC \toadd{($\uparrow$)}& \toadd{HD ($\downarrow$)} & \toadd{95HD ($\downarrow$)} & \toadd{$\mu$D ($\downarrow$)} & Params $\times$10$^3$ & GFLOPs & Train & \toadd{Predict} & \toadd{User}\\
		\hline
		\multirow{6}{*}{\toadd{NL}}&Frangi-NP& \multicolumn{1}{r|}{\toadd{54.16$\pm$8.81}}&\multicolumn{1}{r|}{\toadd{81.04$\pm$18.48}}&\multicolumn{1}{r|}{\toadd{14.78$\pm$13.83}}& \multicolumn{1}{r|}{\toadd{2.47$\pm$2.22}}&\multirow{6}{*}{\toadd{$ \ll 1$}}&\multirow{6}{*}{\toadd{$ \ll 1$}}&\multirow{6}{*}{\toadd{0}}&\multicolumn{1}{r|}{\toadd{25}}&\multicolumn{1}{r|}{\toadd{0}}\\
		&Sato-NP& \multicolumn{1}{r|}{\toadd{55.75$\pm$7.15}}&\multicolumn{1}{r|}{\toadd{78.60$\pm$16.37}}&\multicolumn{1}{r|}{\toadd{11.53$\pm$12.01}}& \multicolumn{1}{r|}{\toadd{2.17$\pm$1.07}}&&&&\multicolumn{1}{r|}{\toadd{25}}&\multicolumn{1}{r|}{\toadd{0}}\\
		&TV-NP & \multicolumn{1}{r|}{\toadd{68.41$\pm$5.01}}&\multicolumn{1}{r|}{\toadd{60.23$\pm$10.08}}&\multicolumn{1}{r|}{\toadd{10.97$\pm$11.72}}& \multicolumn{1}{r|}{\toadd{2.10$\pm$1.00}}&&&&\multicolumn{1}{r|}{\toadd{35}}&\multicolumn{1}{r|}{\toadd{0}}\\ 
		&	Frangi-PP& \multicolumn{1}{r|}{\toadd{68.44$\pm$3.15}}&\multicolumn{1}{r|}{\toadd{\underline{20.60$\pm$10.91}}}&\multicolumn{1}{r|}{\toadd{9.01$\pm$10.38}}& \multicolumn{1}{r|}{{\toadd{2.36$\pm$2.01}}}&&&&\multicolumn{1}{r|}{\toadd{25}}&\multicolumn{1}{r|}{\toadd{25}}\\
		&Sato-PP& \multicolumn{1}{r|}{\toadd{69.01$\pm$3.67}}&\multicolumn{1}{r|}{\toadd{\underline{21.53$\pm$9.11}}}&\multicolumn{1}{r|}{\toadd{8.86$\pm$10.09}}& \multicolumn{1}{r|}{{\toadd{2.10$\pm$1.01}}}&&&&\multicolumn{1}{r|}{\toadd{25}}&\multicolumn{1}{r|}{\toadd{25}}\\
		&TV-PP & \multicolumn{1}{r|}{\toadd{70.74$\pm$3.38}}&\multicolumn{1}{r|}{\toadd{\textbf{20.11$\pm$8.45}}}&\multicolumn{1}{r|}{\toadd{8.31$\pm$8.23}}& \multicolumn{1}{r|}{{\toadd{2.07$\pm$1.02}}}&&&&\multicolumn{1}{r|}{\toadd{35}}&\multicolumn{1}{r|}{\toadd{25}}\\
		\hline
		%
		\multirow{3}{*}{\toadd{FS}} &Vessel 2D-Unet& \multicolumn{1}{r|}{\toadd{\underline{77.66$\pm$4.32}}}&\multicolumn{1}{r|}{\toadd{74.78$\pm$16.73}}&\multicolumn{1}{r|}{\toadd{12.60$\pm$18.16}}& \multicolumn{1}{r|}{\toadd{\underline{0.60$\pm$0.11}}}&\multicolumn{1}{r|}{31.38} & \multicolumn{1}{r|}{15.6} & \multicolumn{1}{r|}{90}& \multirow{3}{*}{\toadd{$<$ 1}}&\multirow{3}{*}{\toadd{327}}\\
		&\toadd{DeepVesselNet}&\multicolumn{1}{r|}{\toadd{\underline{76.13$\pm$5.51}}}&\multicolumn{1}{r|}{\toadd{75.32$\pm$12.94}}&\multicolumn{1}{r|}{\toadd{\underline{4.32$\pm$1.16}}}& \multicolumn{1}{r|}{\toadd{1.65$\pm$0.26}}&\multicolumn{1}{r|}{\toadd{0.05}}&\multicolumn{1}{r|}{NA}&\multicolumn{1}{r|}{\toadd{960}}& & \\
		&2D-WnetSeg& \multicolumn{1}{r|}{\toadd{\underline{76.63$\pm$4.26}}}&\multicolumn{1}{r|}{\toadd{80.69$\pm$23.20}}&\multicolumn{1}{r|}{\toadd{13.15$\pm$19.67}}&\multicolumn{1}{r|}{\toadd{2.13$\pm$2.37}}&\multicolumn{1}{r|}{16.34} & \multicolumn{1}{r|}{25.90} & \multicolumn{1}{r|}{90}&&\\
		
		\hline 
		\multirow{2}{*}{\toadd{LS}} 	&3D-Unet & \multicolumn{1}{r|}{\toadd{68.50$\pm$3.37}}&\multicolumn{1}{r|}{\toadd{76.12$\pm$8.47}}&\multicolumn{1}{r|}{\toadd{15.72$\pm$2.23}}&\multicolumn{1}{r|}{\toadd{2.56$\pm$1.44}}& \multicolumn{1}{r|}{16.21} & \multicolumn{1}{r|}{1669.53} & \multicolumn{1}{r|}{60}  & \multirow{2}{*}{\toadd{$<$ 1}} & \multicolumn{1}{r|}{\toadd{327}}\\
		&\toadd{Pseudo-labeling} & \multicolumn{1}{r|}{\toadd{54.90$\pm$5.86}}&\multicolumn{1}{r|}{\toadd{68.50$\pm$9.58}}&\multicolumn{1}{r|}{\toadd{24.19$\pm$5.25}}& \multicolumn{1}{r|}{\toadd{4.48$\pm$1.67}}&\multicolumn{1}{r|}{\toadd{31.38}} & \multicolumn{1}{r|}{\toadd{15.6}} & \multicolumn{1}{r|}{\toadd{1090}} &  & \multicolumn{1}{r|}{\toadd{0}}\\
		\hline
		&\toadd{PnetCl + K-means} & \multicolumn{1}{r|}{\toadd{64.96$\pm$4.76}}&\multicolumn{1}{r|}{\toadd{65.82$\pm$7.99}}&\multicolumn{1}{r|}{\toadd{16.66$\pm$3.85}}& \multicolumn{1}{r|}{\toadd{2.62$\pm$0.65}} & \multicolumn{1}{r|}{\toadd{0.62}}& \multicolumn{1}{r|}{\toadd{0.993}} & \multicolumn{1}{r|}{\toadd{60}} & \multicolumn{1}{r|}{\toadd{$\sim$1}} & \multirow{2}{*}{\toadd{75.5}}\\
		&Vessel-\verb|CAPTCHA|& \multicolumn{1}{r|}{\toadd{\textbf{79.32$\pm$3.02}}}&\multicolumn{1}{r|}{\toadd{51.70$\pm$5.92}}&\multicolumn{1}{r|}{\toadd{\textbf{4.06$\pm$1.50}}}& \multicolumn{1}{r|}{\toadd{\textbf{0.50$\pm$0.09}}}&\multicolumn{1}{r|}{16.34} & \multicolumn{1}{r|}{25.90} & \multicolumn{1}{r|}{90} & \multicolumn{1}{r|}{\toadd{$<$1}}&\\
		\hline
		\multicolumn{11}{l}{NL: No labels, FS: Fully supervised, LS: Limited supervision, NP: No post-processing, PP: Post-processing, NA: Not Available}\\
		
	\end{tabular}
	\label{tab:summary}
\end{table*}

\begin{figure*}[t]
	\centerline{
		\includegraphics[width=\textwidth]{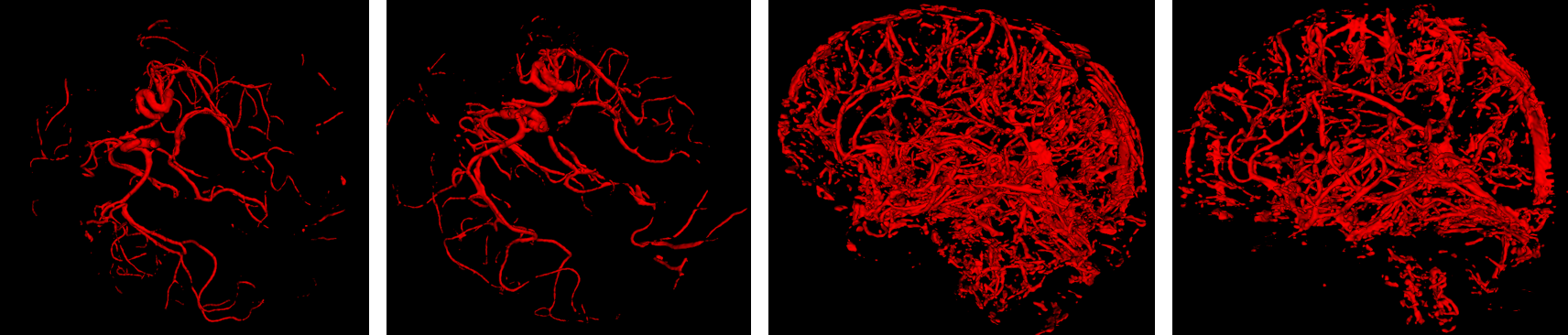} 
	}
	\caption{3D renderings of obtained segmentations in two TOF images (left) and two SWI (right).}
	\label{fig:renderings}
\end{figure*}

\toadd{
	\subsection{Summary}\label{subsec:summary}}
Table~\ref{tab:summary} summarizes the performance of the different baselines compared in this work, along with their computational costs, in terms of model size, FLOPs, training and \toadd{inference time, and user intervention time. We refer to user intervention time as the time to annotate the training set in learning-based approaches, or to post-process the segmentation results for classical methods. It should be noted that for the latter user intervention occurs every time an image is segmented, whereas for learning-based methods this only happens once during training.  In addition to the considered baselines, we include two further methods for reference: the 2D-WnetSeg trained with pixel-wise annotations and the combination of the classifier network with K-means (no segmentation network). Overall, the Vessel-}\verb|CAPTCHA| \toadd{has a performance comparable to the best fully supervised methods \citep{Livne2019}, which avoids post-processing steps, while providing an important speed-up for training data annotation.} 

\vspace{0.3cm}
\section{Discussion and Conclusions}

\toadd{\paragraph{Context and Proposed Solution} D}eep convolutional networks have achieved state-of-the-art performance in many medical image segmentation tasks. However, their success has not been as wide for 3D brain vessel segmentation. This can be explained by two factors. First, deep learning techniques are less performing when the object of interest occupies a small portion of the image, as it is is the case for brain vessels \citep{Livne2019}. Second, manual pixel-wise annotation of vessels is highly time consuming and complex \citep{Moccia2018}. 
In this work, we introduced the Vessel-\verb|CAPTCHA|, an efficient learning framework for vessel annotation and segmentation. The framework formulates the Vessel-\verb|CAPTCHA| annotation scheme, which allows users to annotate a dataset through simple clicks on patches containing vessels, similarly to the commonly used image-CAPTCHAs of web applications \citep{Ahn2004}. \toadd{As such, our work can be considered a multi-instance learning problem where a bag corresponds to an image patch and the instances are the image pixels to be segmented.}

User-provided patch-level tags are used to synthesize pixel-wise \toadd{pseudo-}labels that serve as input to train a 2D patch-based segmentation network. \toadd{In particular,} we use the K-means algorithm to synthesize the pixel-wise \toadd{pseudo-}labels along with the proposed 2D-WnetSeg network, concatenating two 2D-Unets, as backbone architecture.
The use of a 2D patch-based segmentation network instead of more complex end-to-end 3D 
or hybrid architectures, 
is motivated by the need to increase the object-of-interest to image size ratio, as a way to mitigate the reduced performance of deep learning-based methods when the object of interest does not occupy an important portion of the input image. Furthermore, this simplifies the learning process: at a larger scale, the complexity and uniqueness of each brain vessel tree makes it difficult to learn common underlying patterns \citep{Moriconi2019}, whereas, at a local scale, the characteristic patterns of vessels are similar between each other, allowing the network to learn them. Reducing the input size is a common strategy in learning-based vessel segmentation, beyond brain vessel tree segmentation \citep{Kitrungrotsakul2019,kozinski2020}. The lower results obtained by 3D networks validate our choice of a 2D patch-based segmentation network.

\toadd{To further ease the annotation process,} our framework includes a classification network that can label training data without further user effort. This network is trained using the same user-provided patch tags and it allows to classify image patches from unseen images that can be used to enlarge the original training set without the need for further user annotations.

\toadd{\paragraph{Framework Evaluation} W}e  evaluated the proposed framework in terms of its accuracy and required annotation time, \toadd{using a synthetic dataset and two image modalities, TOF and SWI (Fig.~\ref{fig:renderings})}. Our framework achieved performances comparable to those of current state-of-the-art deep learning  approaches for brain vessel segmentation \citep{Livne2019,tetteh2020}, while reducing the annotation burden by \toadd{77\% on average}. \toadd{Moreover, when compared to other approaches subject of limited supervision, our simple yet effective framework demonstrated its superiority.}  
Our promising results, with competitive accuracies and a significant reduction of the user-required effort, should enable the wider use of deep learning techniques for vessel segmentation. 

Our results show that the classifier network not only allows to enlarge the training dataset, but it can act as a \toadd{second opinion to assess the} segmentations. This concept could be further extended to guide a user in the manual correction of a segmentation mask. \toadd{In this work, we used the classification network as an expert. However,} the disagreements between the segmentation and classification network (i.e. 2D-WnetSeg segments a vessel in a patch classified as non-vessel or vice versa) could be used as a measure of uncertainty. Since WnetSeg and PnetCl architectures are significantly different, they extract low-level and high-level features differently. As such, they are complementary to each other: if both agree on a prediction over a patch, the prediction can be considered as one of high confidence, whereas when there is a disagreement the patch can be suggested to the rater for revision. 

\toadd{\paragraph{Limitations and Perspectives} A}lthough our work focuses on the brain vessel tree, we consider that the proposed framework is general enough that it can be easily extended to other vascular structures \citep{Aughwane2019}, other tubular structures with complex networks to annotate \citep{Zuluaga2014}, or different image modalities. However, for some modalities the K-means algorithm used to obtain pixel-wise \toadd{pseudo-}labels can be limited. As an example, the coronary vessel tree imaged with computed tomography angiography is likely to present calcified or lipid plaques that appear as hyper and hypo-intense objects, respectively \citep{zuluaga2011}. In the
current setup, they would be segmented as a vessel (calcified plaques) or the
background (lipid plaques). A natural extension of this work would be to develop novel  self-supervised methods, beyond those studied in this work, which can cope with the characteristics of different vessel/tubular trees and image modalities. 

Our main effort in this work has been directed towards a simplified annotation process and the development of mechanisms that can mitigate the negative effects of `simpler' annotations to achieve performances comparable to the state-of-the-art. Nevertheless, we consider that there are different ways to achieve higher segmentation performance that could be explored. For instance, similarly to what has been proposed by \citep{kozinski2020,Phellan2017}, the annotations could be performed in different image planes. Currently, these are done in the axial plane. In addition, the Vessel-\verb|CAPTCHA| allows for flexible annotations as, for some users, it is simpler to label vessels by following their trajectory. Now, all this information is discarded (see Fig.~\ref{fig:captchas}(e) and (g)), when in some cases it may have relevant content. The challenge here would be to identify when the patch annotations contain relevant information beyond the mere identification of the patch.
Finally, one last limitation of the current framework is related to the selection of the patch grid scheme. While it is convenient to present non-overlapping patches to the user, in some cases, this may degrade the framework's performance. This is particularly true when the grid partition results in the split of vessels, in particular the smaller ones, across two or more patches causing them to lose their characteristic shape. The use of overlapping patches is a straightforward extension of this work that could reduce the number of misclassified vessels.

\section*{Acknowledgments}
ML and MAZ are supported by the French government, through the 3IA Côte d’Azur Investments in the Future project managed by the National Research Agency (ANR) (ANR-19-P3IA-0002). ML is partially funded by the ANR JCJC project Fed-BioMed (19-CE45-0006-01). KL received funding from the Spanish Ministry of Science, Innovation and Universities under grant agreement RTI2018-099898-B-I00 (HeartBrainCom). FP is funded in part by the National Institute for Health Research University College London Hospitals Biomedical Research Centre (NIHR BRC UCLH/UCL High Impact), and by a Non-Clinical Postdoctoral Guarantors of Brain fellowship. 
\bibliographystyle{model2-names.bst}\biboptions{authoryear}
\bibliography{refs}

%

\end{document}